\documentclass[11pt,a4paper]{article}
\usepackage{times,latexsym}
\usepackage{url}
\usepackage[T1]{fontenc}

\usepackage{amssymb}
\usepackage{amsmath}
\usepackage{epsfig}
\usepackage{graphicx}
\usepackage{booktabs}
\usepackage{multirow}
\usepackage{makecell}
\usepackage{caption}
\usepackage{subcaption}
\usepackage{stfloats}
\usepackage{algorithm}
\usepackage{algpseudocode}
\usepackage{arydshln}
\usepackage{xcolor}

\usepackage[acceptedWithA]{tacl2021v1}
\usepackage{xspace,mfirstuc,tabulary}

\newif\iftaclinstructions
\taclinstructionsfalse 
\iftaclinstructions
\renewcommand{\confidential}{}
\renewcommand{\anonsubtext}{(No author info supplied here, for consistency with
TACL-submission anonymization requirements)}
\newcommand{\instr}
\fi

\iftaclpubformat 

\else

\fi

\newcommand{\nls}{\texttt{\textbackslash n}}
\newcommand{\boh}{\texttt{<|start\_header\_id|>}}
\newcommand{\eoh}{\texttt{<|end\_header\_id|>}}
\newcommand{\eos}{\texttt{<|eot\_id|>}}

\newcolumntype{P}[1]{>{\centering\arraybackslash}p{#1}}
\newcolumntype{M}[1]{>{\centering\arraybackslash}m{#1}}

\title{Efficient Tuning of Large Language Models for Knowledge-Grounded Dialogue Generation}

\author{Bo Zhang$^1$ \quad Hui Ma$^2$ \quad Dailin Li$^1$ \quad Jian Ding$^1$ \quad \textbf{Jian Wang}$^1$\Thanks{The corresponding author.} \\
	\textbf{Bo Xu}$^1$ \quad \textbf{HongFei Lin}$^1$ \\
	$^1$Dalian University of Technology, China \quad  $^2$Hefei University of Technology, China \\
    \texttt{\{zhangbo1998,ldlbest,91mr\_ding\}@mail.dlut.edu.cn} \\
    \texttt{huima@hfut.edu.cn, \{wangjian,xubo,hflin\}@dlut.edu.cn} 
}

\date{}

\begin{document}
\maketitle
\begin{abstract}
Large language models (LLMs) demonstrate remarkable text comprehension and generation capabilities but often lack the ability to utilize up-to-date or domain-specific knowledge not included in their training data. To address this gap, we introduce KEDiT, an efficient method for fine-tuning LLMs for knowledge-grounded dialogue generation. KEDiT operates in two main phases: first, it employs an information bottleneck to compress retrieved knowledge into learnable parameters, retaining essential information while minimizing computational overhead. Second, a lightweight knowledge-aware adapter integrates these compressed knowledge vectors into the LLM during fine-tuning, updating less than 2\% of the model parameters. The experimental results on the Wizard of Wikipedia and a newly constructed PubMed-Dialog dataset demonstrate that KEDiT excels in generating contextually relevant and informative responses, outperforming competitive baselines in automatic, LLM-based, and human evaluations. This approach effectively combines the strengths of pretrained LLMs with the adaptability needed for incorporating dynamic knowledge, presenting a scalable solution for fields such as medicine.\footnote{Code and data are available at: \url{https://github.com/zhangbo-nlp/KEDiT}} 
\end{abstract}

\section{Introduction}
The field of natural language processing has undergone a significant transformation recently with the advent of large language models (LLMs) \citep{ref1, ref2, ref3, ref4, ref5}. These models, characterized by their vast number of parameters, have demonstrated remarkable abilities in understanding and generating human-like text, powered by extensive pretraining on diverse and extensive datasets. However, LLMs struggle with tasks that require up-to-date knowledge or domain-specific expertise that was not included in their training datasets \citep{ref6, ref7}. This limitation has led to the exploration of methods to augment LLMs with external knowledge, thereby improving their performance in knowledge-intensive tasks.

One promising approach to address this challenge is retrieval-augmented generation (RAG), a technique that integrates retrieval mechanisms into the generative process of LLMs \citep{ref8, ref9, ref10, ref11, ref12, ref13}. This method allows LLMs to access and utilize external, relevant information dynamically, as they generate responses. Current methods in the RAG system typically employ off-the-shelf LLMs combined with general-purpose retrievers, leveraging the inherent in-context learning capabilities of these language models \citep{ref12, ref14, ref15}. However, this approach encounters limitations when the LLMs are not specifically trained to incorporate retrieved content, particularly in utilizing domain-specific knowledge. These challenges are exacerbated in fields where accurate, specialized information is crucial. In contrast, other researchers have adopted an end-to-end training approach, integrating both LLMs and retrieval mechanisms \citep{ref16, ref17}. This method undoubtedly improves the overall performance of the system by aligning the learning objectives of the model with the retrieval tasks directly. Nonetheless, these extensive training processes are resource-intensive and costly, posing significant challenges for deploying these models in environments that demand up-to-date knowledge integration. Given these challenges, we shift our focus to the generative aspect to enhance knowledge-grounded dialogue generation directly.

Unlike previous approaches to knowledge-grounded dialogue generation \citep{ref18, ref19, ref20}, which often involve a knowledge selection step, we do not consider this step because of its computational expense when it is applied to LLMs, such as with LSR \citep{ref21}. Instead, we propose an efficient method for fine-tuning LLMs for knowledge-grounded dialogue generation, named KEDiT. This method directly utilizes the retrieved knowledge without the selection process and consists of two main stages. \textbf{First}, we employ an information bottleneck method \citep{ref22, ref23} to compress the retrieved knowledge into a set of learnable vectors by maximizing the mutual information between the original knowledge and these compressed vectors. This approach ensures that essential information is retained and reduces the computational complexity of processing extensive knowledge inputs. To further improve this representation, we introduce an alignment loss, refining the compressed vectors to align them with the internal representations of the LLM. \textbf{Second}, we integrate the compressed knowledge representation into the dialogue generation process through a lightweight knowledge-aware adapter (KA-Adapter), which improves the model by inserting small, trainable modules into its architecture. These modules, integrated into both the attention layer and feed-forward layer, selectively fine-tune the model while keeping the majority of its parameters frozen. The adapter employs a gating mechanism as a control channel, regulating how the compressed vectors influence the internal states of the LLM. This design ensures that external knowledge is effectively incorporated without disrupting the pretrained representations. Requiring fine-tuning of less than 2\% of the model parameters, the KA-Adapter balances computational efficiency and high performance in knowledge-grounded dialogue tasks.

To validate the effectiveness of KEDiT, we conduct comprehensive experiments in both open-domain and specialized domain settings. For the open-domain evaluation, we utilize the Wizard of Wikipedia dataset \citep{ref24} to test the ability of models to generate responses grounded in a wide range of knowledge topics. Additionally, to assess performance in specialized domains and with up-to-date information, we create a domain-specific dialogue dataset using GPT-4o, based on the latest research from PubMed.\footnote{\url{https://pubmed.ncbi.nlm.nih.gov/}} The experimental results demonstrate that KEDiT achieves substantial improvements over competitive baselines in automatic, LLM-based, and human evaluations. KEDiT shows superior performance in generating contextually relevant and informative responses and excels in handling domain-specific knowledge. To further validate the practicality of the proposed method, we evaluate KEDiT across various domains and tasks, highlighting its adaptability and robustness in diverse scenarios.

In summary, we present KEDiT, an efficient method for improving knowledge-grounded dialogue generation in large language models. By integrating knowledge-aware components, KEDiT offers a scalable solution to the challenge of incorporating extensive and evolving knowledge bases into dialogue systems without extensive costs. Furthermore, we introduce a new domain-specific dialogue dataset, PubMed-Dialog, which provides a benchmark for assessing the ability of the model to address specialized, up-to-date biomedical information in dialogue generation.

\section{Related Work}
\subsection{Knowledge-Grounded Dialogue}
Knowledge-grounded dialogue systems generate responses that are both contextually appropriate and enriched with relevant information drawn from external knowledge sources. Traditional methods involve pretrained language models such as BART \citep{ref25} and T5 \citep{ref26}, which are fine-tuned on dialogue datasets with explicit knowledge selection and integration \citep{ref24, ref27, ref18, ref28}. These methods typically use a two-step process: selecting relevant knowledge and generating responses on the basis of this information. For example, certain models, such as SPI \citep{ref20}, select the single most relevant knowledge sentence for generation, whereas other models, such as TransIKG \citep{ref19}, integrate multiple knowledge sentences using mechanisms like attention. However, these methods assume that the gold standard knowledge is contained within the available knowledge base, which limits their generalizability. In contrast, \citet{ref29} proposed a weakly supervised learning framework, and \citet{ref30} introduced knowledgeable prefix tuning to inject all relevant knowledge directly into the model, bypassing the need for knowledge selection. However, these innovative methods cannot be applied to LLMs directly because of their specific architectures.

The advent of RAG introduced a new dimension to knowledge-grounded dialogue by combining retrieval mechanisms with LLMs, enabling them to dynamically access external information \citep{ref8, ref9, ref10, ref11, ref12, ref13}. For example, \citet{ref12} show that retrieval-augmented language modeling significantly improves performance by conditioning on relevant documents without modifying the language model. These methods increase the flexibility and applicability of LLMs in knowledge-intensive tasks, but they still struggle to utilize new knowledge in specialized domains efficiently, because they are not specifically trained to incorporate retrieved content. Recent approaches, such as \citet{ref16}, have explored end-to-end training of LLMs with integrated retrieval mechanisms to address these limitations. However, these methods are resource-intensive and challenging to deploy in environments requiring frequent updates.

Our method, KEDiT, addresses these challenges by compressing retrieved knowledge into learnable parameters and integrating them through a lightweight adapter. This approach reduces computational overhead and maintains high performance and adaptability in both open-domain and specialized domain settings.

\subsection{Parameter-Efficient Fine-tuning}
Parameter-efficient fine-tuning techniques adapt LLMs to new tasks with minimal parameter updates, reducing computational costs. These methods can be broadly categorized into two types: adapter-based and prompt-based methods.
 
Adapter-based methods, such as those introduced by \citet{ref31} and extended by others \citep{ref32, ref33, ref34, ref35}, insert additional trainable parameters into the model architecture while keeping most of the original model weights frozen. Among these, low-rank adaptation (LoRA) \citep{ref33} has emerged as a prominent technique that employs a pair of smaller matrices to update the model weights through low-rank decomposition. Prompt-based methods offer another approach to parameter-efficient fine-tuning. These methods, including prompt tuning \citep{ref36}, prefix tuning \citep{ref37}, and P-tuning \citep{ref38, ref39}, prime a frozen pretrained model for a downstream task by including a trainable collection of tokens either in the input embeddings or at every intermediate layer of the model. Additionally, \citet{ref40} combined prefix tuning and LoRA to propose the MAM Adapter, which further refines parameter-efficient fine-tuning by applying modifications to specific hidden states in the model. However, these methods face challenges in knowledge-intensive tasks where specialized information is crucial, often failing to fully capture and utilize the complexity of this knowledge.

Recent parameter-efficient approaches for knowledge-grounded tasks include KnowExpert \citep{xu-etal-2022-retrieval}, which utilizes specialized adapters encoding fixed topic-specific information, and KnowPrefix-Tuning \citep{ref30}, which employs knowledge prefixes to prompt latent information. However, both methods are limited by their reliance on knowledge encoded during pre-training or fixed during training. 

Our proposed KEDiT is specifically designed for knowledge-grounded tasks, focusing on reducing computational overhead while effectively leveraging external dynamic knowledge. Furthermore, we innovatively adapt the prompt-based method into an adapter-based approach with a gating mechanism, ensuring seamless integration of compressed knowledge vectors into the LLM for improved dialogue generation.

\section{Method}
\subsection{Task Statement and Model Overview}
\begin{figure*}
    \centering
    \begin{subfigure}[b]{0.39\textwidth}
        \centering
        \includegraphics[width=\textwidth]{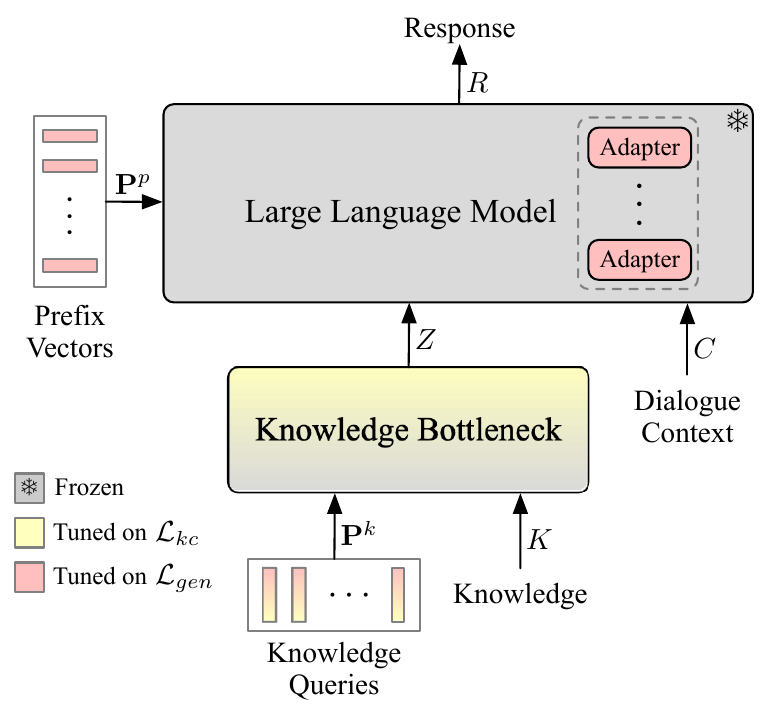}
        \caption{Method Overview}
        \label{fig:method_overview}
    \end{subfigure}
    \hfill
    \begin{subfigure}[b]{0.28\textwidth}
        \centering
        \includegraphics[width=\textwidth]{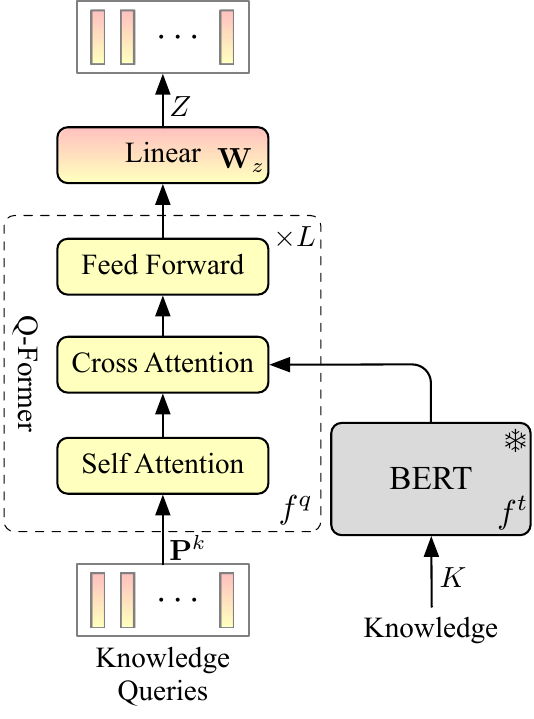}
        \caption{Knowledge Bottleneck}
        \label{fig:method_kb}
    \end{subfigure}
    \hfill
    \begin{subfigure}[b]{0.28\textwidth}
        \centering
        \includegraphics[width=\textwidth]{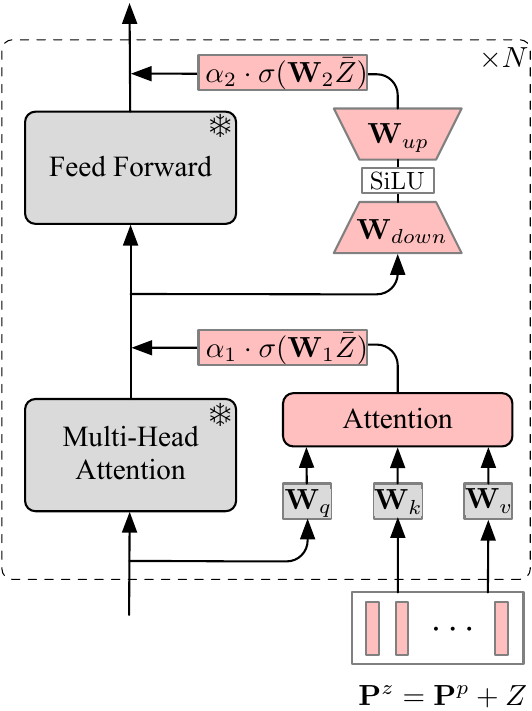}
        \caption{Large Language Model}
        \label{fig:method_llm}
    \end{subfigure}
    \hfill
    \caption{(a) Overall architecture of KEDiT, showing the flow from the input knowledge and dialogue context to the generated response. (b) Detailed structure of the knowledge bottleneck module, showing how BERT and the Q-Former compress knowledge into a compact representation through multi-head attention and feed-forward layers. (c) Integration of compressed knowledge into the large language model using KA-Adapter, detailing KA-Attn and KA-FFN. For simplicity, we omit adding \& norm layers to the diagrams. Yellow indicates tuning in the knowledge compression step, pink indicates tuning in the dialogue generation step, and gray represents frozen modules.}
    \label{fig:method}
\end{figure*}

Given a dialogue context $C$ and a set of retrieved knowledge pieces $K = \{k_1, k_2, \ldots, k_n\}$, the task is to generate a response $R$ that is both contextually appropriate and enriched with the provided knowledge. Formally, we aim to maximize the conditional probability $p(R|C, K)$. However, using $K$ directly can be computationally expensive for LLMs.
To address this issue, we propose KEDiT, as shown in Figure \ref{fig:method_overview}, which comprises two main components: a knowledge bottleneck module and a frozen LLM enhanced with KA-Adapter. The knowledge bottleneck module distills essential information from $K$ into a compact representation $Z$, formalized by $p_\phi(Z|K)$, where $Z$ is a set of learnable vectors. The LLM then incorporates $Z$ into the dialogue generation process via the KA-Adapter, represented by $p_\theta(R|C, Z)$. Thus, our revised objective becomes:
\begin{equation}
	p(R|C, K) \approx p_\theta(R|C, Z)p_\phi(Z|K).
\end{equation}

In summary, our approach follows a two-step strategy: knowledge compression to distill $K$ into $Z$ and knowledge integration to efficiently utilize $Z$ within the dialogue generation process. The training and inference procedures are described in the following sections, with an overview provided in Appendix~\ref{appendix:algorithms}.

\subsection{Knowledge Compression via an Information Bottleneck}
Integrating external knowledge into LLMs presents significant challenges due to the vast length of retrieved information, which often contains irrelevant details and increases computational costs. To address this issue, we propose a knowledge compression mechanism based on the information bottleneck principle. This method distills essential information into a fixed-size, learnable representation, balancing relevance and efficiency for seamless integration with LLMs. Our approach trains a knowledge bottleneck module, combining BERT \citep{ref41} and the Q-Former \citep{ref22}, that is optimized using an information bottleneck objective on a large-scale knowledge dataset $\mathcal{D}_k$.

\paragraph{Knowledge Encoding}
We utilize a pretrained BERT to encode the knowledge $K$ into feature representations $f^t(K)$, specifically using the last hidden states. These features are then processed by the Q-Former, which compresses them into a compact representation $Z$. As shown in Figure \ref{fig:method_kb}, the Q-Former consists of $L$ blocks, each including a self-attention layer, followed by a cross-attention layer that incorporates $f^t(K)$, and a feed-forward network. Each block is formally defined as:
\begin{equation}
	\label{eq:z}
	\mathbf{Z_\ell} = f_\ell^q(\mathbf{Z}_{\ell-1}, f^t(K)) \in \mathbb{R}^{m \times d_1},
\end{equation}
for $\ell = 1, \ldots, L$, with the initial input $\mathbf{Z}_0 = \mathbf{P}^k$, where $\mathbf{P}^k \in \mathbb{R}^{m \times d_1}$ represents $m$ learnable vectors referred to as knowledge queries. These vectors are randomly initialized and interact with $f^t(K)$, which enables them to absorb the semantic and contextual information from $K$. The concept of knowledge queries is inspired by prompt tuning \cite{ref36}. They serve as trainable vectors designed to capture essential information from $K$ and integrate it efficiently into the LLM.

After processing through all $L$ blocks, we obtain the final representation $Z = \mathbf{W}_z\mathbf{Z}_L \in \mathbb{R}^{m \times d_2}$, where $\mathbf{W}_z$ is a learnable projection matrix that maps the representation to the dimension required by the LLM.

\paragraph{Mutual Information Optimization}
To ensure that the compressed $Z$ retains the most essential information from $K$ and can be fully utilized by the LLM, we maximize their mutual information $I(K;Z)$ via the LLM. A common approach for achieving this is to maximize a variational lower bound \citep{ref42} on the mutual information, which is expressed as:
\begin{equation}
	I(K;Z) \ge \mathbb{E}_{p(K,Z)}\log q(K|Z) + H(K).
\end{equation}

However, in our setup, $Z$ is a set of learnable vectors rather than latent variables sampled from a specific distribution. Therefore, we adapt this approach by introducing an auxiliary model $q_\psi(K|Z)$ to reconstruct $K$ from $Z$. This model is parameterized by the LLM, thereby ensuring that $Z$ can be effectively utilized by the LLM in the dialogue generation process. Consequently, our optimization objective simplifies to:
\begin{equation}
	\label{eq:recon}
	\mathcal{L}_{recon} = - \mathbb{E}_{p(K)p_\phi(Z|K)}\log q_\psi(K|Z),
\end{equation}
where $p_\phi(Z|K)$ is modeled by the knowledge bottleneck. This formulation is similar to the variational lower bound approach, where $p(K,Z)$ is factorized as $p(K) \times p_\phi(Z|K)$ and the entropy term $H(K)$, which is constant with respect to the model parameters, can be omitted during optimization. Although this adaptation diverges from traditional variational approaches, we find that it works well in practice.

\paragraph{Alignment Loss}
While mutual information optimization ensures that $Z$ retains essential information from $K$, it does not guarantee that $Z$ is readily interpretable by the LLM. To further refine the compressed knowledge vectors $Z$ and align them closely with the LLM's internal representations, we introduce an alignment loss. This loss is designed to ensure that the compressed knowledge vectors are easily interpretable and utilizable by the LLM. Specifically, alignment loss ensures that the structure of the compressed vectors $Z$ produced by the knowledge bottleneck is compatible with the vectors $\hat{Z}$ reconstructed by the LLM from the original knowledge, where $\hat{Z}$ corresponds to the final hidden states of the LLM. We define this loss as the mean squared error between the vectors $Z \sim p_\phi(Z|K)$ and the vectors $\hat{Z} \sim q_\psi(Z|K)$:
\begin{equation}
	\label{eq:align}
	\mathcal{L}_{align} = \frac{1}{m} \sum_{i=1}^{m} (Z_i - \hat{Z}_i)^2.
\end{equation}

\paragraph{Training Objective}
The overall objective for knowledge compression combines mutual information optimization and alignment loss:
\begin{equation}
	\label{eq:kc}
	\mathcal{L}_{kc} = \mathcal{L}_{recon} + \beta \cdot \mathcal{L}_{align},
\end{equation}
where $\beta$ is a hyperparameter that balances the contribution of the alignment loss. During this training phase, the parameters of the LLM and BERT are kept frozen, whereas only the parameters of the Q-Former are trained.

By minimizing this objective, we ensure that the compressed knowledge vectors $Z$ capture essential information from $K$ effectively while being well-aligned with the internal representations of the LLM. This compressed representation is then used in the dialogue generation process, as described in the subsequent sections.

\subsection{Knowledge Integration into Dialogue Generation}
After the compressed knowledge representation $Z$ is obtained, the next step is to integrate this knowledge into the LLM to improve dialogue generation while maintaining computational efficiency. Directly fine-tuning the LLM to incorporate external knowledge is both resource-intensive and risks disrupting the pretrained representations of the model. To address this issue, we design the KA-Adapter, a lightweight module that integrates external knowledge by inserting small, trainable components into the LLM layers. This approach preserves the pretrained capabilities of the model and enables efficient fine-tuning on a knowledge-grounded dialogue dataset $\mathcal{D}_d$. As depicted in Figure \ref{fig:method_llm}, this adapter consists of two main components: the knowledge-aware attention mechanism (KA-Attn) and the knowledge-aware feed-forward network (KA-FFN).

\paragraph{Knowledge-Aware Attention Mechanism}
KA-Attn improves the standard self-attention mechanism of LLMs by incorporating $Z$. Inspired by an alternative view of prefix tuning \citep{ref40}, we transform prompt-based prefix tuning into an adapter-based method, adapting the attention mechanism as follows:
\begin{align}
	\label{eq:ka-attn}
	&\mathbf{h_1} \gets \mathbf{h_1} + \alpha_1 \cdot \sigma(\mathbf{W}_1\bar{Z}) \cdot \triangle \mathbf{h_1}, \notag \\
	&\triangle \mathbf{h_1} = \mathrm{Softmax}(\mathbf{W}_q\mathbf{x}(\mathbf{W}_k\mathbf{P}^z)^T) \mathbf{W}_v\mathbf{P}^z,
\end{align}
where $\mathbf{P}^z = \mathbf{P}^p + Z$, $\mathbf{P}^p \in \mathbb{R}^{m \times d_2}$ represents $m$ learnable prefix vectors, $\mathbf{x}$ is the input feature, and $\sigma$ denotes the $sigmoid$ activation. $\mathbf{W}_q$, $\mathbf{W}_k$, and $\mathbf{W}_v$ are the query, key, and value matrices, respectively, of the original attention mechanism. The function $\alpha_1 \cdot \sigma(\mathbf{W}_1\bar{Z})$ acts as a gating mechanism, allowing the model to dynamically adjust its focus on the adaptive change on the basis of the mean value of the compressed vectors $Z$.

\paragraph{Knowledge-Aware Feed-Forward Network}
Similarly, the KA-FFN is designed to integrate the compressed knowledge vectors into the standard FFN layer of LLMs. This mechanism draws inspiration from LoRA and a scaled parallel adapter \citep{ref40} and is adapted to better integrate external knowledge as follows:
\begin{align}
	\label{eq:ka-ffn}
	&\mathbf{h_2} \gets \mathbf{h_2} + \alpha_2 \cdot \sigma(\mathbf{W}_2\bar{Z}) \cdot \triangle \mathbf{h_2}, \notag \\
	&\triangle \mathbf{h_2} = \mathbf{W}_{up} \cdot \mathrm{SiLU}(\mathbf{W}_{down} \mathbf{h_1}),
\end{align}
where $\mathbf{W}_{down}$ and $\mathbf{W}_{up}$ are matrices that transform $\mathbf{h_1}$ into a lower and then back to a higher dimension, effectively performing a bottleneck operation. The term $\alpha_2 \cdot \sigma(\mathbf{W}_2\bar{Z})$ functions similarly to the gating mechanism in KA-Attn, modulating the impact of $\triangle \mathbf{h_2}$ on the FFN's output.

\paragraph{Generating Response}
To generate the response $R$, the LLM predicts each token $R_t$ sequentially, conditioned on the dialogue context $C$, the compressed representation $Z$, and the previously generated tokens $R_{<t}$. At each step, the input embeddings of $C$ and $R_{<t}$ are concatenated with $Z$ and then passed into the LLM's stacked $N$ layers with the KA-Adapter. The final hidden state $\mathbf{h}_{R_t}$ of the current token is then projected into the vocabulary space to compute the next-token probabilities:
\begin{equation}
	\label{eq:log}
	p_\theta(R_t|R_{<t}, C, Z) = \mathrm{Softmax}(\mathbf{W}_o \mathbf{h}_{R_t}),
\end{equation}
where $\mathbf{W}_o$ is the output projection matrix.

\paragraph{Training Objective}
The training objective for this stage focuses on ensuring that the LLM generates contextually appropriate and knowledge-enriched responses. We achieve this by minimizing the negative log-likelihood of the target response tokens:
\begin{equation}
	\label{eq:gen}
	\mathcal{L}_{gen} = -\sum_{t} \log p_\theta(R_t|R_{<t}, C, Z),
\end{equation}
where $(C, K, R) \in \mathcal{D}_d$ and $Z \sim p_\phi(Z|K)$. In this step, only the parameters of the KA-Adapter and the projection matrix $\mathbf{W}_z$ are tuned, whereas the parameters of the LLM and the remainder of the knowledge bottleneck remain frozen.

\section{Experiments}
\subsection{Datasets}
To evaluate the effectiveness of KEDiT, we employ three datasets. The Wikipedia dataset serves as the knowledge corpus ($\mathcal{D}_k$) for knowledge compression training. The Wizard of Wikipedia and PubMed-Dialog datasets are used for training and evaluating dialogue generation ($\mathcal{D}_d$).

\paragraph{Wikipedia}
For the knowledge compression phase, we use the English version of the Wikipedia dataset.\footnote{\url{https://huggingface.co/datasets/wikimedia/wikipedia}} This dataset is built from Wikipedia dumps and includes cleaned articles. We further process this dataset by splitting articles into paragraphs, selecting up to 500 words of content per article, and discarding articles with fewer than 50 words. This process results in a high-quality dataset of 6 million text chunks, which is suitable for training our knowledge bottleneck model. 

\paragraph{Wizard of Wikipedia}
This dataset \citep{ref24} is used to evaluate the model performance in open-domain dialogue generation. The dataset contains 22.3k dialogues with 100.8k turns, where the agent provides informative responses grounded in Wikipedia knowledge. The dataset is divided into training, validation, and test sets, with the validation and test sets further split into seen and unseen categories. The seen category includes topics present in the training set, whereas the unseen category consists of dialogues with topics not encountered during training. In our experiments, instead of using the predefined knowledge sentences provided in the dataset, we retrieve three relevant knowledge pieces from the provided knowledge topic articles on the basis of the dialogue context using TF-IDF, as described in \citep{ref24}. Notably, only 80\% of the dialogue turns contain the predefined gold knowledge sentence among the retrieved topics. Although this may slightly impact performance, it better simulates real-world scenarios.

\paragraph{PubMed-Dialog}
To evaluate the model performance in specialized domains, we construct the PubMed-Dialog dataset using GPT-4o. Specifically, we use a data filtering methodology similar to that used in PubMedQA \citep{ref43} to select relevant latest research articles from the PubMed database. Next, we design a prompt to instruct GPT-4o to generate multi-turn dialogues on the basis of the knowledge from the corresponding abstracts of these articles. This prompt is crafted to simulate natural conversations about biomedical topics, ensuring that the dialogues cover a range of aspects related to the topics discussed in the abstract. This provides a comprehensive benchmark for assessing the ability of the model to address specialized, up-to-date knowledge in dialogue generation. Notably, we directly use the abstract as knowledge of the corresponding dialogue context. To ensure the quality and faithfulness of the generated dialogues and minimize hallucinations, where content may deviate from the source information or contain inaccuracies, we implement a multi-iterative validation process. This process involves three rounds of evaluation and regeneration, iteratively refining the dialogues until they meet the required standards of consistency and factual correctness. Following this process, we obtain a dataset of 10.9k dialogues, with each dialogue containing an average of 4.36 turns. The dataset is divided into training, validation, and test sets, with 80\% for training, 10\% for validation, and 10\% for testing. Details on the prompt design and the multi-iterative validation process are provided in Appendix~\ref{appendix:prompt_example}.

\subsection{Experimental Setup}
\paragraph{Baseline Models}
To evaluate the performance of KEDiT, we compare it against several baseline models grouped into three categories. First, the standard language models, including BART-Large \citep{ref25} and Llama-3-8B \citep{ref4} using LoRA (\textsc{Llama$_{lora}$}), generate dialogue responses without external knowledge, which serve as basic benchmarks. Second, traditional knowledge-grounded models, such as TransIKG \citep{ref19} and SPI \citep{ref20}, improve dialogue generation by selecting and incorporating predefined knowledge sentences from fixed sources in the original dataset. Third, the retrieved knowledge-augmented models, such as KAT-TSLF \citep{ref29}, Llama-3-8B using KnowPrefix-Tuning \citep{ref30} (\textsc{Llama$_{kpt}$}), and Llama-3-8B using retrieved knowledge for zero-shot generation (\textsc{Llama$_{rag}$}), utilize knowledge pieces retrieved on the basis of the dialogue history using the same TF-IDF method as in our experiments. We apply the same data processing methods within each category. All baselines are fine-tuned on the target dialogue dataset for fair comparison.

\paragraph{Implementation Details}
In our implementation of KEDiT, we utilize Llama-3-8B as the frozen LLM. Both the BERT encoder and the Q-Former are initialized with weights from BERT$_{base}$ \citep{ref41}.
During the knowledge compression phase, the knowledge bottleneck is trained on $\mathcal{D}_k$ using the AdamW optimizer with a batch size of 64 over 1 epoch. The training configuration includes a learning rate of 2e-4, 10,000 warmup steps, a weight decay of 0.05, and $\beta$ is 0.5. The parameter $m$ is 16 to balance the trade-off between the expressiveness of the compressed knowledge representation and computational efficiency. The impact of different values of $m$ is discussed in Section~\ref{sec:impact_know}. The models $q_\psi(Z|K)$ and $q_\psi(K|Z)$ are generated using instructions detailed in Appendix~\ref{appendix:instruction_template}. 
In the dialogue generation phase, the KA-Adapter is fine-tuned on $\mathcal{D}_d$ using the AdamW optimizer with a learning rate of 1e-4 and cosine decay. Training is conducted over 3 epochs with a batch size of 32, and $\alpha_1$ and $\alpha_2$ are 2 and 4, respectively. The generation model $p_\theta(R|C, Z)$ uses instruction templates specified in Appendix~\ref{appendix:instruction_template} to concatenate $C$ and $Z$. We concatenate all knowledge snippets collectively and input them into BERT for encoding.

\begin{table*}[ht!]
	\centering
	\textsc{\resizebox{.96\textwidth}{!}{
	\begin{tabular}{lcccccccccr}
		\toprule
		\multirow{2}{*}{Model} & \multicolumn{3}{c}{WoW Seen} & \multicolumn{3}{c}{WoW Unseen} & \multicolumn{3}{c}{PubMed-Dialog} & \multirow{2}{*}{\makecell{Tuned \\ Params}} \\
		\cmidrule(lr){2-4} \cmidrule(lr){5-7} \cmidrule(lr){8-10}
		 & F1 & BLEU & ROUGE & F1 & BLEU & ROUGE & F1 & BLEU & ROUGE & \\
		\midrule
		BART$_{large}$ & 20.54 & 12.10 & 14.73 & 18.38 & 10.53 & 12.78 & 32.30 & 21.17 & 23.53 & 406M \\
		Llama$_{lora}$ & 20.45 & 12.37 & 14.95 & 20.26 & 12.14 & 14.71 & 35.77 & \underline{23.87} & 26.38 & 168M \\
		TransIKG & 21.31 & 12.77 & \underline{16.61} & 19.40 & 11.71 & 15.08 & - & - & - & 194M \\
		SPI-uniform & \underline{21.82} & \underline{13.05} & 16.43 & \textbf{21.14} & \textbf{12.80} & \underline{15.89} & - & - & - & 141M \\
		KAT-TSLF & 20.50 & 12.45 & 15.13 & 19.60 & 11.97 & 14.30 & \underline{36.16} & 23.52 & \underline{27.05} & 198M \\
		Llama$_{rag}$ & 17.07 & 8.69 & 11.61 & 17.41 & 8.84 & 11.94 & 34.40 & 20.80 & 26.05 & 0M \\
		Llama$_{kpt}$ & 19.28 & 10.80 & 13.99 & 18.28 & 9.93 & 13.07 & 29.43 & 18.66 & 23.98 & 214M \\
		\midrule
		KEDiT & \textbf{22.45}$^\star$ & \textbf{13.87}$^\star$ & \textbf{17.24}$^\star$ & \underline{21.05} & \underline{12.63} & \textbf{15.94} & \textbf{38.63}$^\star$ & \textbf{25.84}$^\star$ & \textbf{28.91}$^\star$ & 140M \\
		- kc & 21.05 & 12.60 & 15.81 & 20.08 & 11.76 & 15.00 & 37.04 & 24.01 & 27.43 & 140M \\
		- $\mathcal{L}_{align}$ \textnormal{in} kc & 22.01 & 13.16 & 16.92 & 20.64 & 12.03 & 15.81 & 37.93 & 25.28 & 28.40 & 140M \\
		- ka-adapter & 18.41 & 9.71 & 12.35 & 17.72 & 9.57 & 12.17 & 33.42 & 20.86 & 25.12 & 3M \\
		- ka-attn & 22.06 & 13.38 & 16.86 & 20.79 & 12.16 & 15.52 & 38.10 & 25.65 & 28.46 & 138M \\
		- ka-ffn & 20.37 & 11.77 & 14.47 & 19.13 & 10.77 & 13.97 & 35.52 & 22.93 & 26.20 & 5M \\
		\bottomrule
	\end{tabular}
	}}
	\caption{Automatic evaluation results on Wizard of Wikipedia (WoW) and PubMed-Dialog test sets. The best results are shown in \textbf{bold}, and the second-best results are \underline{underlined}. The table also includes ablation experiments showing the performance effect of removing key components of KEDiT. Significant improvements over the best baseline are marked by $^\star$ (one-sample t-test, $p < 0.05$).}
	\label{tab:automatic}
\end{table*}

\subsection{Evaluation Methods}
\paragraph{Automatic Evaluation}
To quantify the performance of KEDiT, we utilize the following metrics: (1) BLEU, which measures the precision of n-grams in the generated responses compared with reference responses, with the final score being the average of BLEU-1, BLEU-2, BLEU-3, and BLEU-4; (2) ROUGE, which evaluates the recall of n-grams, focusing on ROUGE-1, ROUGE-2, and ROUGE-L, with the final score being the average of these three metrics; and (3) F1 score, specifically the unigram F1 score, which is the harmonic mean of precision and recall for unigrams.

\paragraph{LLM-Based Evaluation}
Inspired by the evaluation framework in \citet{ref44}, we use GPT-4o as an advanced judge to assess the quality of the responses of our model. We employ two key methods: pairwise comparison, where GPT-4o is presented with a user query and two responses (one response from a baseline and one response from KEDiT) and selects the better response on the basis of relevance, informativeness, accuracy, and coherence; and multi-response grading, where GPT-4o is presented with responses from several baselines and KEDiT simultaneously and assigns scores from 1 to 5 for relevance, informativeness, and fluency for each response. This method has demonstrated high agreement with human evaluations, exceeding 80\% alignment, comparable to human-human agreement levels \citep{ref44}. For a detailed evaluation, we randomly sample 500 examples each from the test seen and test unseen sets of the Wizard of Wikipedia dataset, and 1000 examples from the PubMed-Dialog test set. Specific prompts for these evaluations are provided in Appendix~\ref{appendix:prompt_template}.

\paragraph{Human Evaluation}
To complement the LLM-based evaluation and validate its reliability, we conduct a human evaluation as an additional assessment of the model performance. Specifically, we randomly sample 20\% of the data from the LLM-based evaluation sets: 100 examples each from the test seen and test unseen sets of the Wizard of Wikipedia dataset, and 200 examples from the PubMed-Dialog test set, totaling 400 samples. We recruited 12 graduate students with expertise in natural language processing and bioinformatics to serve as evaluators. These evaluators are divided into four groups of three, with each group independently evaluating 100 samples. Evaluators assess responses on the basis of relevance, informativeness, and fluency using a 1-to-5 Likert scale, following the same criteria as the LLM-based evaluation. To ensure consistency and reduce subjective variance, evaluators receive detailed guidelines and participate in a calibration session before scoring. The evaluation is conducted under a double-blind setup to eliminate bias. Additionally, we compute interrater agreement using Cohen's Kappa \citep{ref45} to analyze the consistency between evaluators and their alignment with the LLM-based judge.

\subsection{Main Results}

\subsubsection{Automatic Evaluation Results}
The automatic evaluation results of KEDiT, compared with those of various baseline models on the Wizard of Wikipedia and PubMed-Dialog datasets, are summarized in Table \ref{tab:automatic}. KEDiT consistently outperforms all baseline models across most evaluation metrics in both datasets. In the test seen set, KEDiT achieves the highest scores, indicating superior performance in generating contextually relevant and informative responses. For the test unseen set, KEDiT demonstrates robust generalization capabilities, although it slightly lags behind SPI in F1 and BLEU. This is primarily because SPI uses predefined gold knowledge and a training method similar to LSR, which is effective but computationally expensive. However, KEDiT is designed to be applicable to a broader range of scenarios and does not rely on predefined gold knowledge. This flexibility allows KEDiT to perform well even in environments where predefined knowledge is not available.

In the PubMed-Dialog dataset, KEDiT significantly outperforms all the baselines, highlighting its exceptional ability to generate accurate and informative responses in specialized domains. Traditional knowledge-grounded models are not suitable for these scenarios because they depend on predefined knowledge, which is not available in the PubMed-Dialog dataset. The existing retrieved knowledge-augmented methods do not perform well on this dataset, likely because they are not optimized to effectively utilize the retrieved specialized knowledge. The knowledge bottleneck mechanism compresses and integrates knowledge more effectively, ensuring that KEDiT can utilize the knowledge during dialogue generation.

KEDiT also demonstrates efficiency, requiring fine-tuning of only 140M parameters, less than 2\% of the LLM's 8B parameters. These results underscore the effectiveness of KEDiT in compressing and integrating knowledge, leading to substantial improvements while maintaining efficiency.

\begin{table*}[ht!]
	\centering
	\textsc{\resizebox{\textwidth}{!}{
	\begin{tabular}{clcccccc}
		\toprule
		\multirow{2}{*}{Method} & \multirow{2}{*}{Model} & \multicolumn{3}{c}{Wizard of Wikipedia} & \multicolumn{3}{c}{PubMed-Dialog}  \\
		\cmidrule(lr){3-5} \cmidrule(lr){6-8}
		&  & relevance & informativeness & fluency & relevance & informativeness & fluency  \\
		\midrule
		\multirow{4}{*}{\makecell{LLM-based\\ Evaluation}}
		& Llama$_{lora}$ & 3.87 & 2.61 & 4.46 & 4.46 & 3.42 & 4.86 \\
		& SPI-uniform & 3.29 & 2.49 & 4.25 & - & - & -  \\
		& KAT-TSLF & 3.16 & 2.31 & 4.18 & 3.88 & 3.34 & 4.45  \\
		& KEDiT & \textbf{3.92} & \textbf{2.72}$^\star$ & \textbf{4.59}$^\star$ & \textbf{4.85}$^\star$ & \textbf{3.82}$^\star$ & \textbf{4.89}  \\
		\midrule
		\multirow{4}{*}{\makecell{Human\\ Evaluation}}
		& Llama$_{lora}$ & 3.91 & 3.11 & 4.52 & 4.50 & 3.83 & 4.80 \\
		& SPI-uniform & 3.76 & 3.04 & 4.28 & - & - & -  \\
		& KAT-TSLF & 3.50 & 3.00 & 4.21 & 4.02 & 3.58 & 4.57  \\
		& KEDiT & \textbf{4.09}$^\star$ & \textbf{3.36}$^\star$ & \textbf{4.60} & \textbf{4.72}$^\star$ & \textbf{4.03}$^\star$ & \textbf{4.82}  \\
		\bottomrule
	\end{tabular}
	}}
	\caption{Multi-response grading evaluation results on Wizard of Wikipedia and PubMed-Dialog test sets. Significant improvements over the best baseline are marked by $^\star$ (independent t-test, $p < 0.05$).}
	\label{tab:grading}
\end{table*}

\subsubsection{LLM-Based Evaluation Results}
\begin{figure}
    \centering
    \begin{subfigure}[b]{0.23\textwidth}
        \centering
        \includegraphics[width=\textwidth]{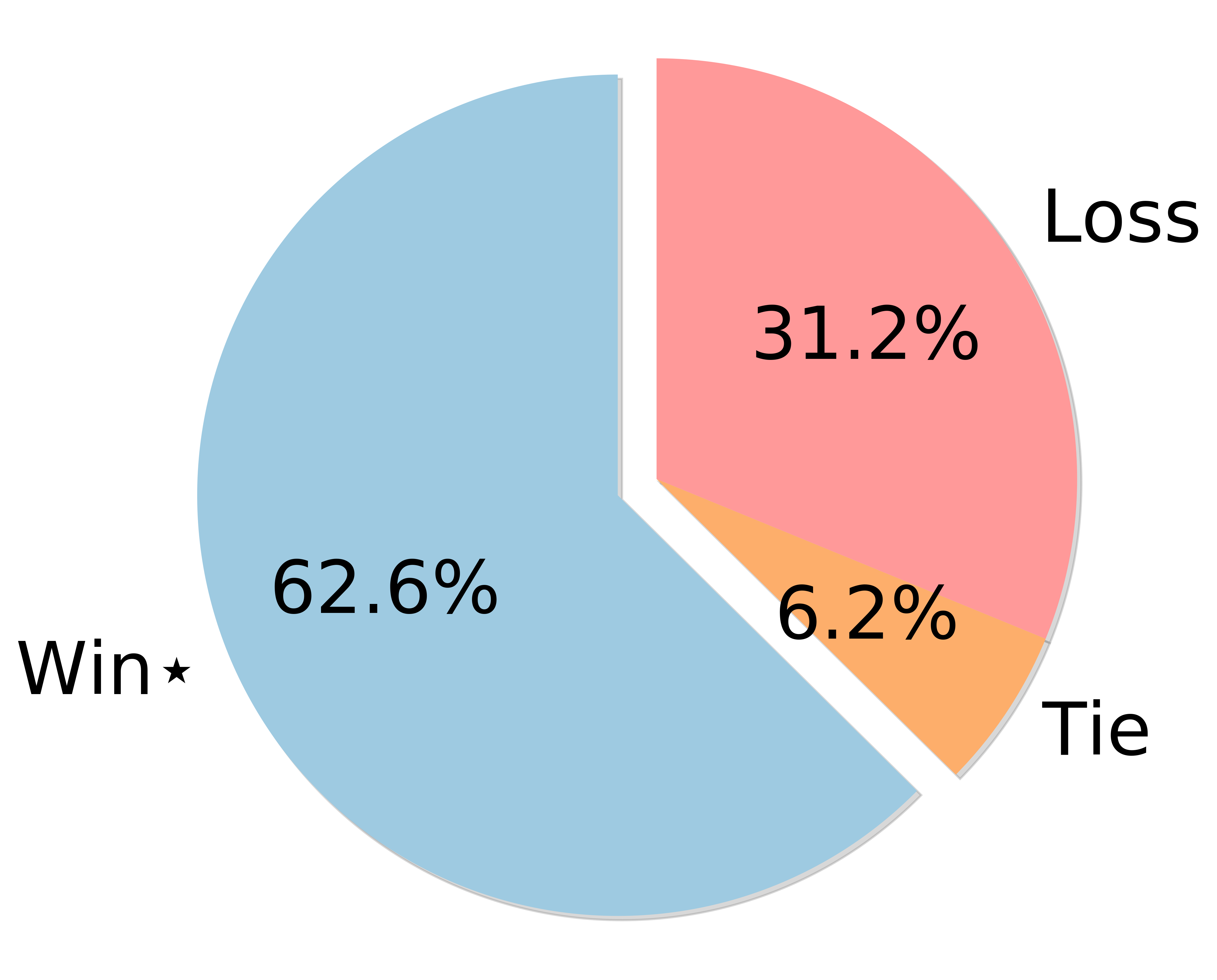}
        \caption{Wizard of Wikipedia}
        \label{fig:wow_pairwise}
    \end{subfigure}
    \hfill
    \begin{subfigure}[b]{0.24\textwidth}
        \centering
        \includegraphics[width=\textwidth]{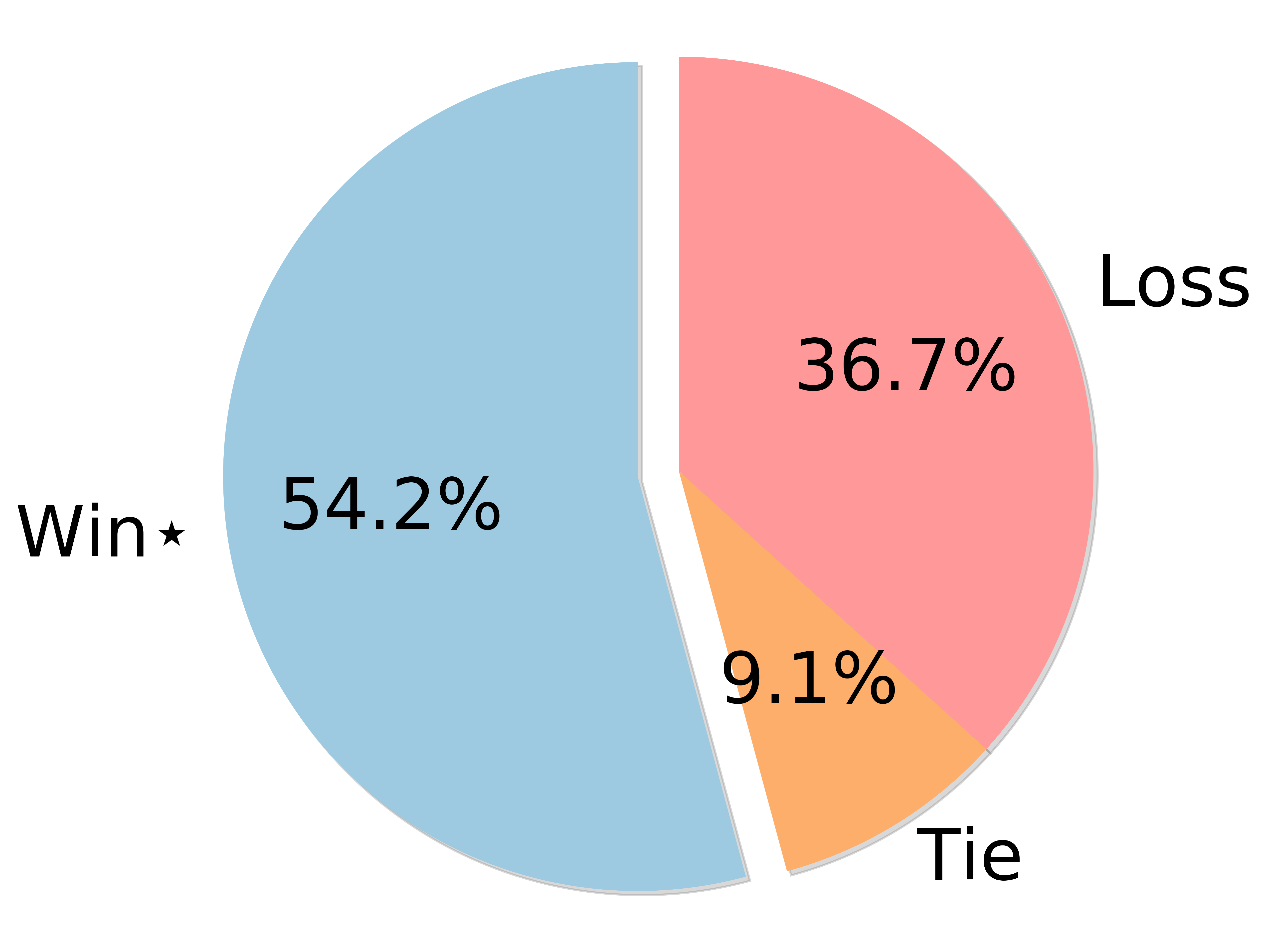}
        \caption{PubMed-Dialog}
        \label{fig:pubmed_pairwise}
    \end{subfigure}   
	\caption{Pairwise comparison results of KEDiT against baseline models, showing win, tie, and loss rates. The comparisons are against SPI on the Wizard of Wikipedia test sets and Llama$_{lora}$ on the PubMed-Dialog test set. Significant improvements are marked with $^\star$ (binomial test, $p < 0.01$).}
    \label{fig:pairwise}
\end{figure}

In the pairwise comparisons, we evaluate KEDiT against the best-performing baseline models in automatic evaluation metrics, namely SPI for the Wizard of Wikipedia dataset and Llama$_{lora}$ for the PubMed-Dialog dataset. As shown in Figure \ref{fig:pairwise}, KEDiT significantly outperforms SPI on the Wizard of Wikipedia dataset, achieving a higher win rate and demonstrating a superior ability to generate relevant and informative responses. Similarly, on the PubMed-Dialog dataset, KEDiT achieves a notable win rate over Llama$_{lora}$, underscoring its effectiveness in incorporating domain-specific knowledge into the dialogue generation process.

In the multi-response grading evaluation, we compare KEDiT with the best-performing models from three baseline categories on the basis of automatic evaluation. As detailed in the upper part of Table \ref{tab:grading}, compared with these baselines, KEDiT consistently achieves higher scores for relevance, informativeness, and fluency. Interestingly, although SPI scores highly on automatic metrics, it performs worse than does Llama$_{lora}$ in LLM-based evaluations. This discrepancy is likely due to Llama's robust language generation capabilities, which produce more coherent and contextually appropriate responses. SPI's reliance on predefined knowledge may limit its adaptability, resulting in less natural dialogue flow. KEDiT, based on the Llama model, combines the strengths of both informativeness and fluency, achieving superior results in both automatic and LLM-based evaluations.

\begin{figure*}[ht!]
    \centering
    \begin{subfigure}[b]{0.41\textwidth}
    	\centering
	    \includegraphics[width=\textwidth]{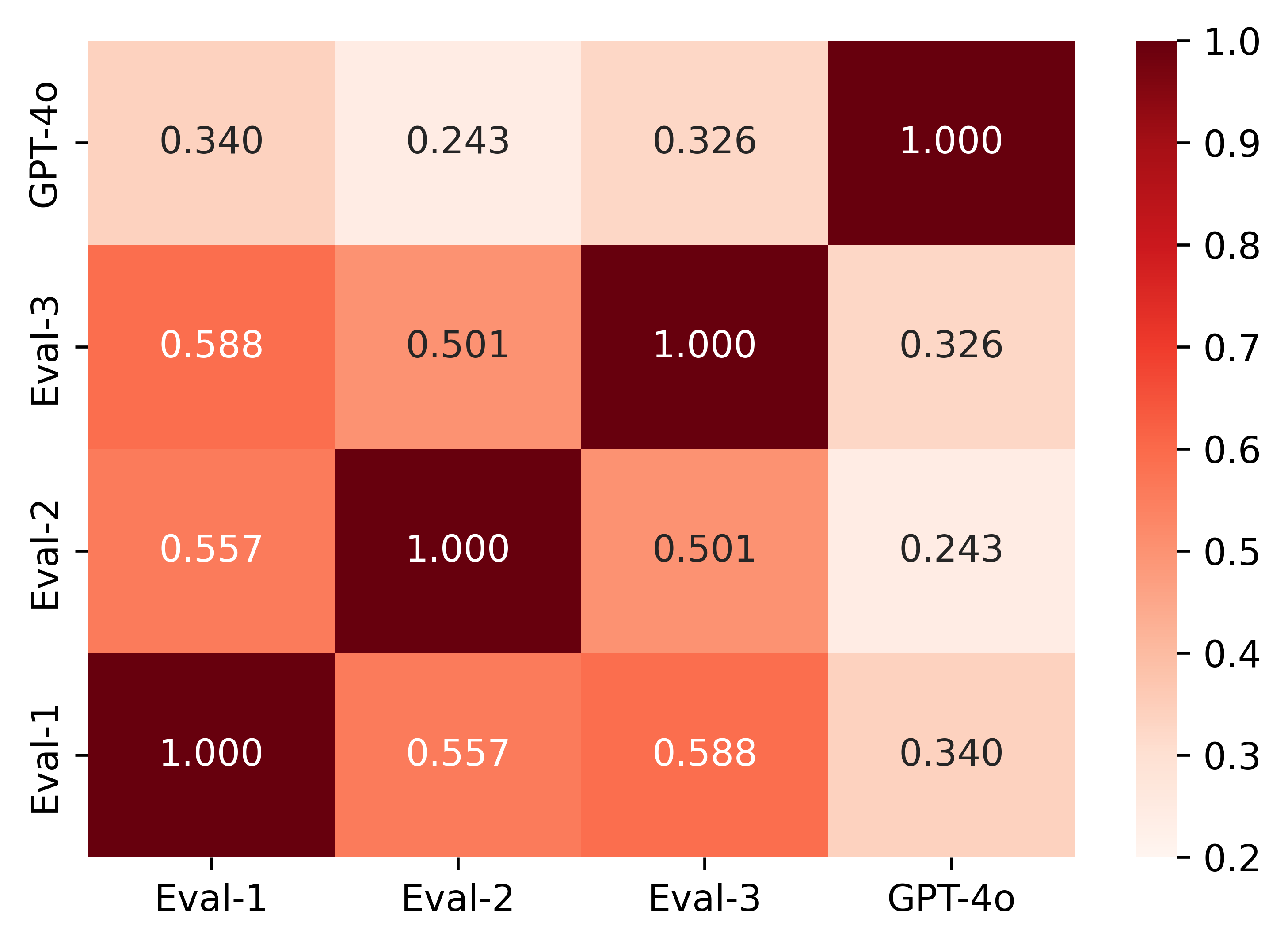}
        \caption{Wizard of Wikipedia (Seen)}
    \end{subfigure}
    \hfil
    \begin{subfigure}[b]{0.4\textwidth}
        \centering
        \includegraphics[width=\textwidth]{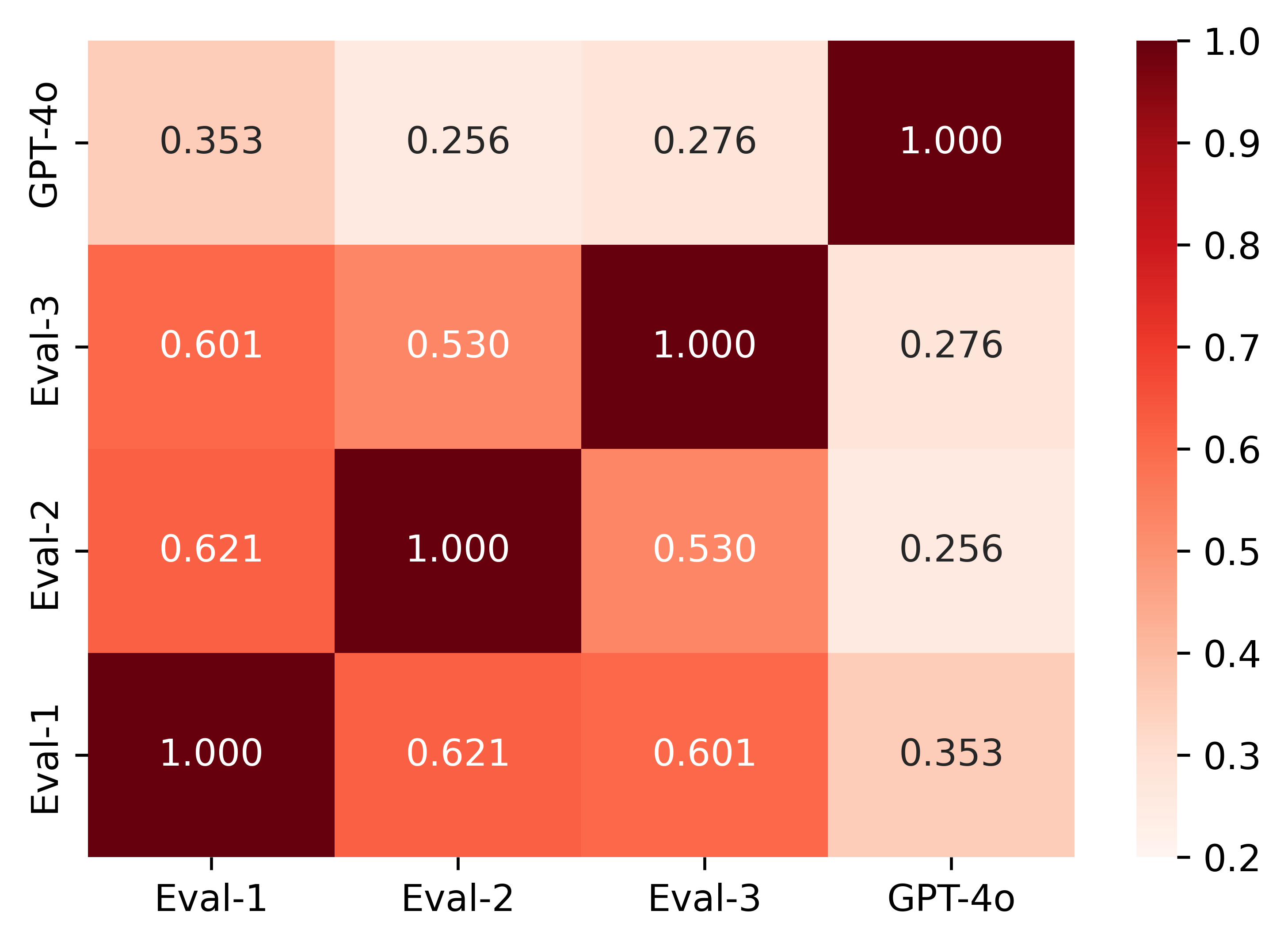}
        \caption{Wizard of Wikipedia (Unseen)}
    \end{subfigure}
    \begin{subfigure}[b]{0.4\textwidth}
		\centering
	    \includegraphics[width=\textwidth]{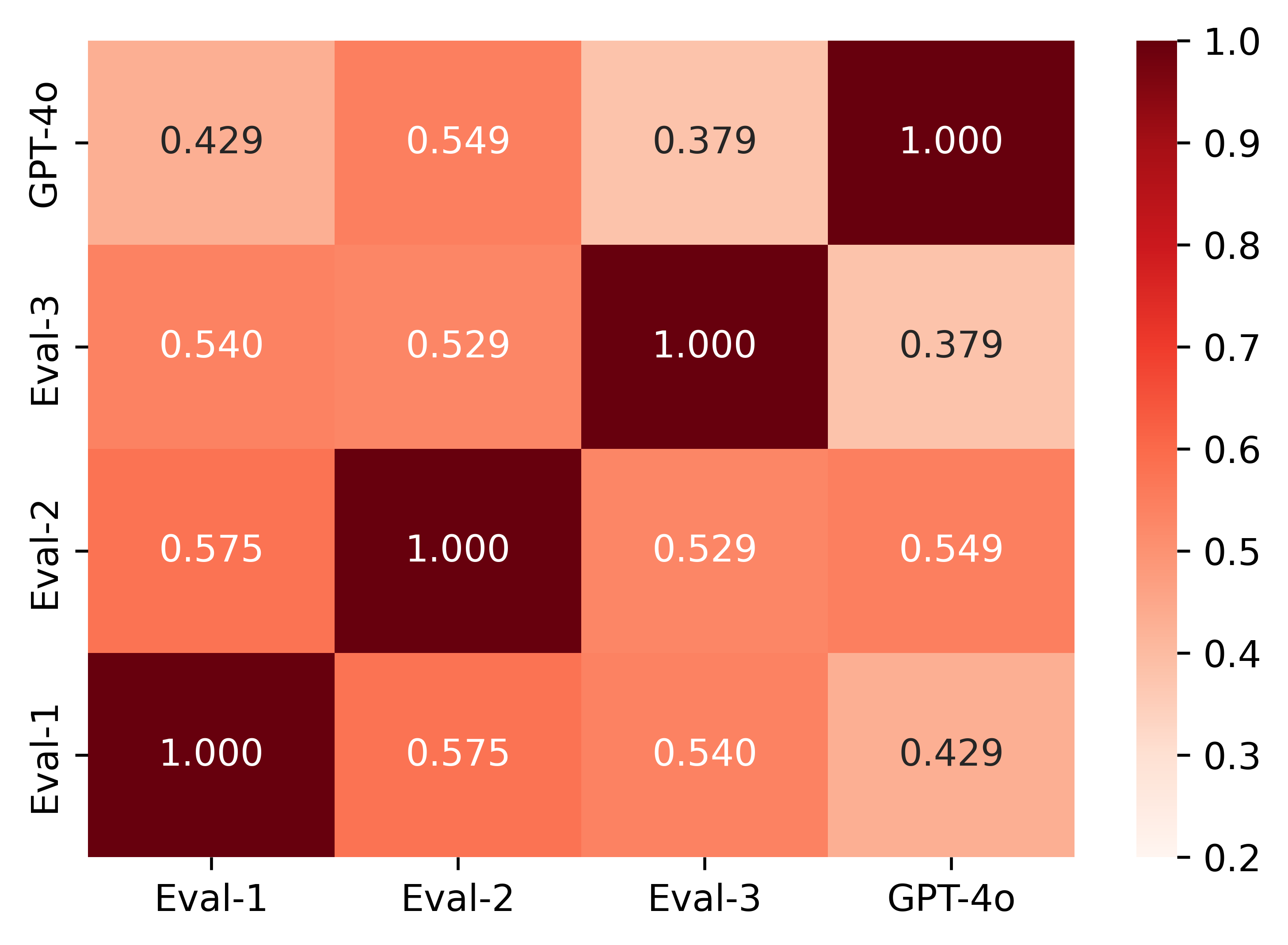}
        \caption{PubMed-Dialog (Set 1)}
    \end{subfigure}
    \hfil
    \begin{subfigure}[b]{0.4\textwidth}
        \centering
        \includegraphics[width=\textwidth]{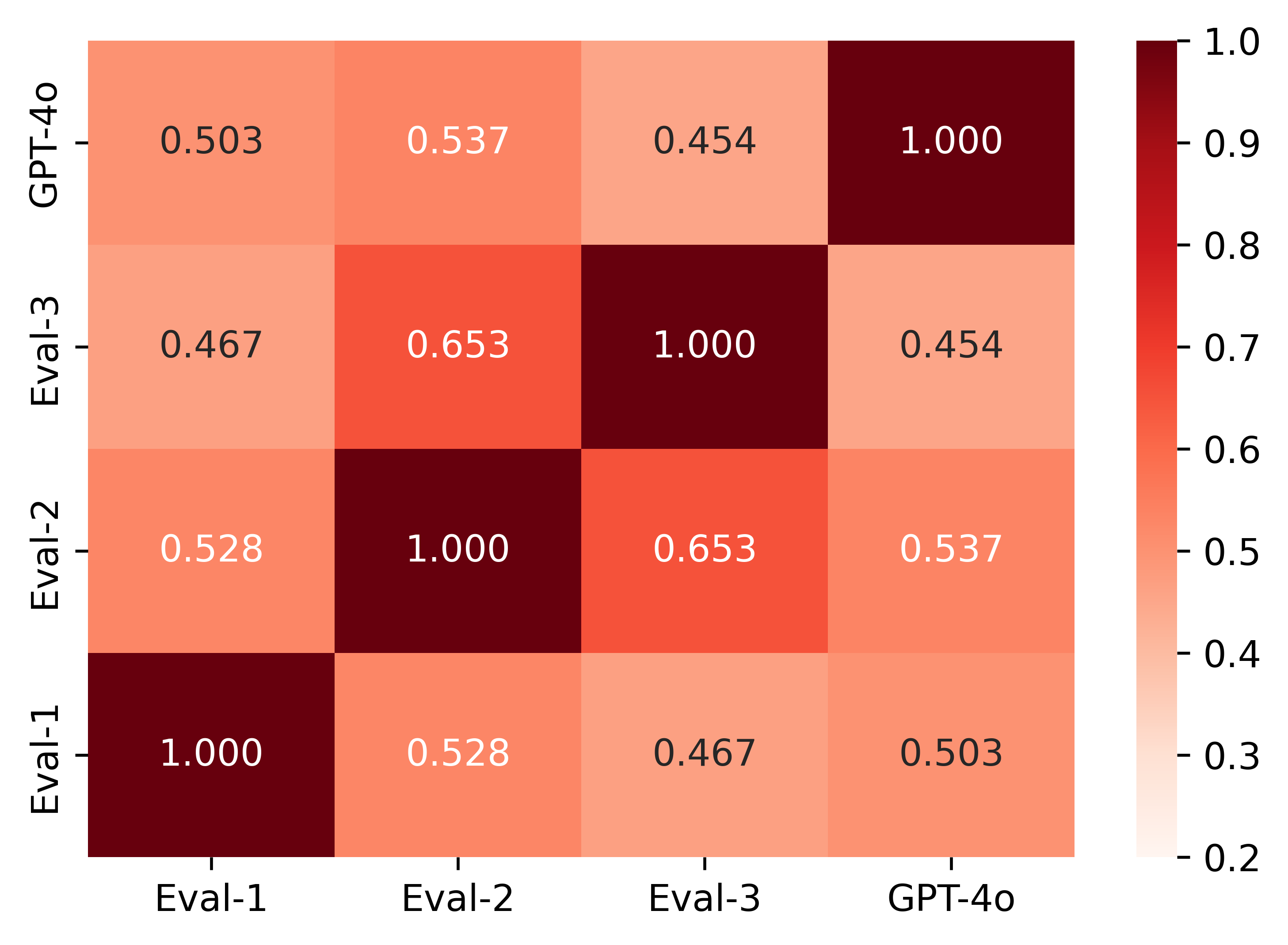}
        \caption{PubMed-Dialog (Set 2)}
    \end{subfigure}   
	\caption{Heatmaps of Cohen's Kappa coefficient matrix showing the agreement between evaluators (abbreviated as Eval-1, Eval-2, and Eval-3) and GPT-4o. Higher coefficients indicate greater agreement.}
    \label{fig:kappa}
\end{figure*}

\subsubsection{Human Evaluation Results}
The human evaluation results are summarized in the lower part of Table \ref{tab:grading}. Consistent with the LLM-based evaluation, KEDiT consistently outperforms the baseline models across all the metrics in both datasets. Notably, the improvements in relevance and informativeness scores are statistically significant in both datasets, highlighting KEDiT's superior ability to generate responses that are not only contextually appropriate but also rich in information. The lack of significant improvement in fluency is because both the strong baseline and KEDiT already perform exceptionally well. Additionally, we observe that compared with the LLM-based evaluation, human scores tend to be slightly more tempered, with less pronounced score differences between models for the same sample. 

To quantify the agreement between human evaluators and the LLM-based judge, we compute Cohen's Kappa coefficients, which are visualized in Figure~\ref{fig:kappa}. The interrater agreement between human evaluators, with Kappa coefficients between human evaluators generally above 0.5 in most cases, indicates moderate to substantial agreement. Similarly, the agreement between human evaluators and GPT-4o also spans from fair to moderate, with Kappa coefficients generally above 0.24 and reaching 0.55 in some cases. However, lower Kappa scores are observed primarily in the Wizard of Wikipedia dataset, likely due to the shorter responses in this dataset, because GPT-4o tends to assign lower informativeness scores to brief responses. This level of agreement indicates that GPT-4o's evaluations are reasonably consistent with human judgments, which supports the reliability of GPT-4o as a judge.

\subsection{Analysis}
\subsubsection{Ablation Study}
To assess the impact of each component in KEDiT, we perform an ablation study by removing individual modules and assessing performance on the Wizard of Wikipedia and PubMed-Dialog datasets. The lower part of Table \ref{tab:automatic} shows that removing any module results in decreased performance across automatic evaluation metrics, highlighting the importance of each component. 

Specifically, removing the knowledge-aware adapter results in the most significant decrease. This is likely because, without it, the knowledge queries alone cannot effectively integrate compressed knowledge into dialogue generation. The KA-Adapter provides specialized mechanisms to incorporate this knowledge, ensuring its effective utilization. Removing KA-Attn or KA-FFN also results in decreased performance, which further confirms the critical role of these submodules in facilitating knowledge integration. Removing the knowledge compression module (\textsc{kc}) also causes a substantial decrease, which highlights its importance in efficiently distilling essential information from retrieved knowledge. Excluding alignment loss ($\mathcal{L}_{align}$) results in a smaller but noticeable degradation, which indicates the necessity of aligning compressed knowledge with the internal representations of the LLM.

\begin{figure*}[ht!]
	\centering
	\includegraphics[width=\textwidth]{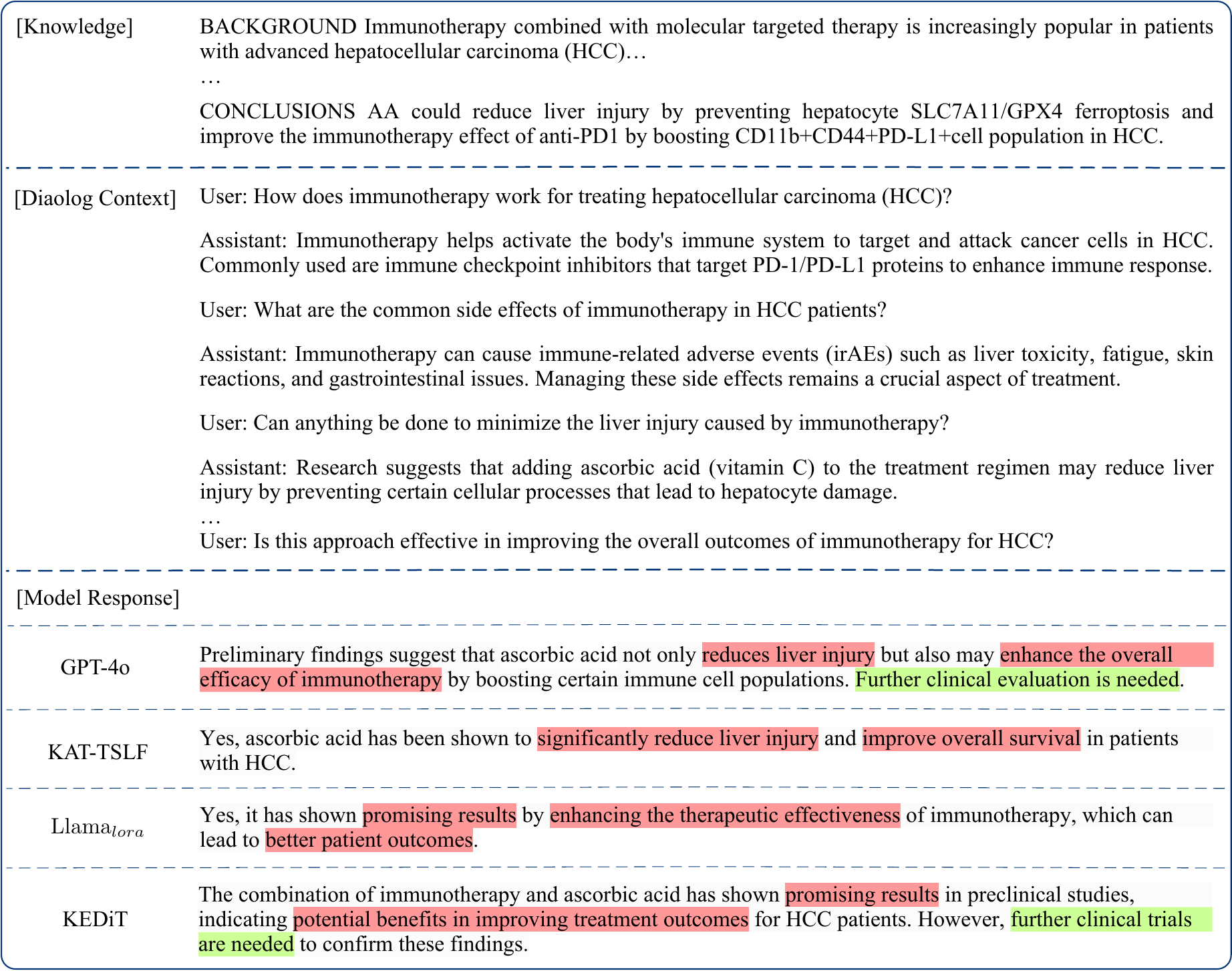}
	\caption{Case study from the PubMed-Dialog test set comparing KEDiT with several baselines. The dialogue shows user questions and assistant responses on immunotherapy for HCC.}
	\label{fig:case}
\end{figure*}

\subsubsection{Case Study}
Figure \ref{fig:case} shows a dialogue from the PubMed-Dialog test set where the user queries about immunotherapy for HCC. When the user asks about the effectiveness of combining ascorbic acid (AA) with immunotherapy for improving overall outcomes, KEDiT responds that this combination has shown promising results in preclinical studies, indicating potential benefits but emphasizing the need for further clinical trials. This response balances optimism with caution, providing a comprehensive and realistic assessment. In contrast, KAT-TSLF states that ascorbic acid significantly reduces liver injury and improves overall survival, which lacks a nuanced perspective on the need for further validation. Llama$_{lora}$ mentions the approach's promise in enhancing therapeutic effectiveness but does not address the necessity for additional clinical trials, making its response less thorough. Compared with the response generated by GPT-4o, which highlights the potential benefits and calls for further evaluation, KEDiT similarly emphasizes the need for additional clinical trials, offering a balanced and detailed response.

\subsubsection{Cross-Model Generalization Analysis}
To evaluate the generalizability and robustness of KEDiT across different large language models, we integrate our framework with two additional state-of-the-art open-source LLMs: Qwen2.5 \citep{yang2024qwen2} and Mistral-v0.3 \citep{jiang2023mistral7b}, each with 7B parameters. To ensure a fair and consistent evaluation, we conduct experiments using the same training and inference pipelines as those used for Llama-3-8B. Additionally, we fine-tune each LLM using LoRA to establish baseline performances. As shown in Table \ref{tab:llm_comparison}, KEDiT consistently outperforms LoRA-based fine-tuning across Llama-3, Qwen2.5, and Mistral-v0.3 on both the Wizard of Wikipedia and PubMed-Dialog test sets. It is worth noting that the results for Qwen$_{kedit}$ and Mistral$_{kedit}$ are slightly lower compared to Llama$_{kedit}$, primarily because Llama$_{kedit}$ was specifically fine-tuned with hyperparameter optimization, while the same parameters were directly applied to Qwen$_{kedit}$ and Mistral$_{kedit}$. Nevertheless, the advantage of KEDiT over LoRA remains significant. Compared to LoRA, KEDiT's knowledge compression and adapter-based integration enable more efficient and targeted utilization of external knowledge, leading to higher F1 scores in both open-domain and specialized-domain tasks. These results highlight the scalability and adaptability of KEDiT, showing that its lightweight yet powerful mechanism for knowledge integration can generalize well across different LLM families.

\begin{table}
	\centering
	\begin{sc}
	\resizebox{\columnwidth}{!}{
	\begin{tabular}{lccc}
		\toprule
		Model & WoW Seen & WoW Unseen & PMD \\
		\midrule
		Llama$_{lora}$ & 20.45 & 20.26 & 35.77 \\
		Mistral$_{lora}$  & 20.09 & 19.73 &  35.52 \\
		Qwen$_{lora}$ & 20.51 & 20.11 & 35.92 \\
		\midrule
		Llama$_{kedit}$ & 22.45 & 21.05 & 38.63 \\
		Mistral$_{kedit}$ & 21.94 & 20.76 & 37.69 \\
		Qwen$_{kedit}$ & 22.15 & 20.94 & 38.31 \\
		\bottomrule
	\end{tabular}
	}
	\end{sc}
	\caption{F1 scores comparing LoRA-based fine-tuning and KEDiT-enhanced models across different LLMs.}
	\label{tab:llm_comparison}
\end{table}

\subsubsection{Impact of Knowledge Queries}
\label{sec:impact_know}
We conduct experiments to assess the effect of varying the number of knowledge queries ($m$) on KEDiT's performance, with values of $m$ set at 2, 4, 8, 16, and 32. Our results, summarized in Table \ref{tab:prompt}, reveal a positive correlation between increasing $m$ and model performance. However, while higher $m$ values generally improve performance, the gains beyond setting $m$ to 16 are minimal. This diminishing return may be due to the model reaching a saturation point where additional queries contribute little new information. Moreover, setting $m$ to 16 allows the model to be efficiently trained on a consumer-grade GPU with 24 GB of memory, whereas setting $m$ to 32 exceeds this capacity, making training impractical on such hardware.

\begin{table}
	\centering
	\begin{sc}
	\resizebox{.5\textwidth}{!}{
	\begin{tabular}{rccc}
		\toprule
		$m$ & WoW Seen & WoW Unseen & PubMed-Dialog \\
		\midrule
		2 & 21.32 & 20.58 & 37.06 \\
		4 & 21.85 & 20.48 & 37.64 \\
		8 & 22.01 & 20.97 & 38.37 \\
		16 & \textbf{22.45} & 21.05 & 38.63 \\
		32 & 22.33 & \textbf{21.15} & \textbf{38.69} \\
		\bottomrule
	\end{tabular}
	}
	\end{sc}
	\caption{F1 scores for different numbers of knowledge queries on the Wizard of Wikipedia and PubMed-Dialog test sets.}
	\label{tab:prompt}
\end{table}

\begin{figure*}[ht!]
    \includegraphics[width=\textwidth]{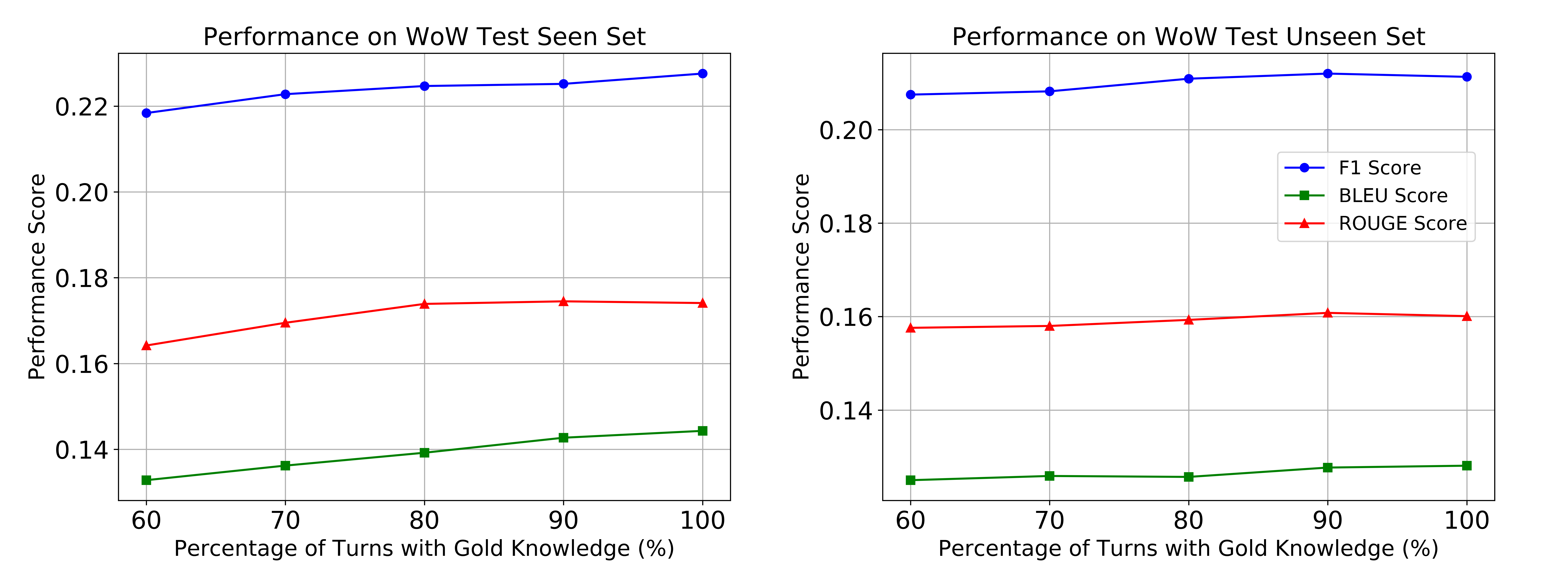}
	\caption{Performance of KEDiT with different percentages of gold knowledge retrieval.}
   	\label{fig:retrieval}
\end{figure*}

\subsubsection{Impact of Retrieval Performance}
To evaluate the impact of retrieval performance on KEDiT, we conduct experiments by varying the percentage of dialogue turns containing the predefined gold knowledge sentence in the Wizard of Wikipedia dataset. Since KEDiT itself does not include a retrieval component, this indirect approach helps assess how the accuracy of retrieved knowledge affects dialogue generation. We conduct tests with percentages of 60\%, 70\%, 80\%, 90\% and 100\% and evaluate the performance of KEDiT under these conditions.
As shown in Figure \ref{fig:retrieval}, the performance of KEDiT generally improves as the percentage of turns with gold knowledge increases. However, the overall differences are not very large, indicating that KEDiT maintains a robust level of performance even with less-than-perfect knowledge retrieval. Notably, the performance gains are more pronounced in the seen set than in the unseen set. This finding clearly suggests that the model benefits more from accurate knowledge retrieval when dealing with familiar topics. In contrast, the unseen set involves new topics, which limits the degree of improvement even with high retrieval accuracy. While the performance of KEDiT in unfamiliar domains is somewhat constrained, its lightweight nature allows for quick adaptation to new domains with minimal retraining.

\begin{table*}[ht]
\centering
\textsc{\resizebox{\textwidth}{!}{
\begin{tabular}{l@{\hskip 1em}ccccccc}
	\toprule
	& \multicolumn{1}{c}{Fact Check.}  & \multicolumn{1}{c}{Entity Linking} & \multicolumn{1}{c}{Slot Filling} & \multicolumn{4}{c}{Open Domain QA}  \\
	Model & \textbf{FEVER}  & \textbf{AY2} & \textbf{zsRE}  & \textbf{NQ} & \textbf{HotpotQA} & \textbf{TriviaQA} & \textbf{ELI5} \\
	\midrule
	\multicolumn{4}{c|}{Accuracy} & \multicolumn{3}{c|}{Exact Match} & \multicolumn{1}{c}{ROUGE-L}  \\
	\midrule
	BART & 79.80 & 82.66 & 5.29 & 13.18 & 14.46 & 18.40 & 20.04 \\
	Llama$_{lora}$ & 87.22 & 90.16 & 28.11 & 40.36 & 27.69 & 43.65 & 20.63 \\
	Llama$_{rag}$ & \textcolor{gray}{90.53} & \textcolor{gray}{57.29} & \textcolor{gray}{66.00} & \textcolor{gray}{54.53} & \textcolor{gray}{43.60} & \textcolor{gray}{86.97} & 18.17 \\
	\midrule
	KEDiT & 88.83 & 91.70 & 32.85 & 44.50 & 28.38 & 62.11 & 22.19 \\
	\bottomrule
\end{tabular}
}}
\caption{Performance on knowledge intensive tasks (dev sets). }
\label{tab:kilt}
\end{table*}

\subsubsection{Evaluation on the KILT Benchmark}
To assess the adaptability and robustness of KEDiT across diverse domains and tasks, we conduct additional experiments using the KILT benchmark \citep{ref46}. This benchmark is a widely recognized framework for evaluating models on a variety of knowledge-intensive language tasks, encompassing tasks such as fact-checking, entity linking, and open-domain question answering. We evaluate KEDiT on the following tasks included in KILT: FEVER \citep{ref47}, AIDA CoNLL-YAGO (AY2; \citealp{ref48}), Zero Shot RE \citep{ref49}, Natural Questions (NQ; \citealp{ref50}), HotpotQA \citep{ref51}, TriviaQA \citep{ref52}, and ELI5 \citep{ref53}. Metrics include Accuracy, Exact Match (EM), and ROUGE-L, following KILT's standard evaluation procedures. We leverage the off-the-shelf Contriever-MS MARCO \citep{ref54} to retrieve three relevant documents for each input.

Table~\ref{tab:kilt} summarizes the results across all evaluated tasks. In evaluating Llama$_{rag}$, we observe significant discrepancies between its predictions and ground-truth labels on most tasks except ELI5. When directly computing evaluation metrics, these discrepancies result in extremely low scores, often approaching zero. To address this issue and make the comparison more meaningful, we adopt a relaxed evaluation approach: if the ground-truth label appears in the predictions of Llama$_{rag}$, the prediction is considered correct. This adjustment has significantly inflated metric values for Llama$_{rag}$. Despite these adjustments for Llama$_{rag}$, KEDiT consistently outperforms baselines, demonstrating superior adaptability and robustness across knowledge-intensive tasks. The ELI5 task stands out in the evaluation as it features long-form answers, which align well with KEDiT's strengths. This suggests that KEDiT's design makes it particularly well-suited for tasks requiring detailed and extended responses, a characteristic often observed in dialogue scenarios.

\section{Discussion}

\subsection{Potential Biases in PubMed-Dialog}
While the PubMed-Dialog dataset is constructed to enhance knowledge-grounded dialogue in the biomedical domain, it may inherit biases from its data sources. First, PubMed primarily consists of peer-reviewed research articles, which may introduce selection bias by over-representing academic perspectives while underrepresenting clinical or patient viewpoints. Second, since the dialogues are synthesized using GPT-4o, there is potential for stylistic and terminological biases that could affect model generalization to non-expert users. Moreover, models from the GPT family (or those pretrained on GPT-generated data) may exhibit inflated performance due to inherent similarities in language patterns and knowledge representation. This could lead to an unfair advantage for such models compared to others not exposed to GPT-generated content during training. Finally, the dataset captures knowledge at a fixed point in time, meaning that newer medical discoveries may not be adequately reflected. While these biases are inevitable to some extent, we employ iterative validation processes to minimize their impact. Additionally, we plan to explore alternative evaluation strategies in future work, such as cross-utilizing different LLMs for both generation and evaluation, to further mitigate potential overfitting to GPT-generated text.

\subsection{Limitations of Automatic Metrics in Dialogue Evaluation}
While BLEU and ROUGE remain widely used for the automatic evaluation of text generation models, they have notable limitations in dialogue evaluation. These metrics primarily focus on n-gram overlaps with reference responses, making them insufficient for assessing contextual coherence, factual correctness, and diversity of generated responses. Prior works \citep{liu-etal-2016-evaluate, novikova-etal-2017-need} have demonstrated that such surface-based metrics correlate poorly with human judgments in dialogue settings. Given these shortcomings, we complement our evaluation with LLM-based scoring and human assessments to better capture the conversational quality and informativeness of the generated responses.

\section{Conclusion}
This paper presents KEDiT, a novel approach for improving knowledge-grounded dialogue generation in LLMs. KEDiT effectively addresses the limitations of LLMs in utilizing up-to-date and domain-specific knowledge by compressing external knowledge into learnable parameters and integrating them using a lightweight adapter. To support this evaluation, we create the PubMed-Dialog dataset, which provides a benchmark for assessing the ability of the model to address specialized biomedical knowledge. Our extensive experiments on the Wizard of Wikipedia and PubMed-Dialog datasets demonstrate that, compared with existing methods, KEDiT significantly improves the contextual relevance and informativeness of generated responses. Further analysis highlights KEDiT's robust performance even with varying retrieval accuracy, maintaining high levels of contextual relevance and informativeness. Although KEDiT shows slightly reduced performance on unseen domains, its design allows for efficient retraining and deployment in environments requiring frequent updates to knowledge. 
Future work will focus on enabling dynamic knowledge updates, enhancing generalization to unseen domains, and integrating multimodal knowledge for greater adaptability and practicality.

\section*{Acknowledgments}
This work is supported by the National Natural Science Foundation of China (No. 62376051) and the Anhui Provincial Natural Science Foundation (2408085QF188).
We thank the anonymous TACL reviewers and the action editor, Xiaojun Wan, for their insightful feedback. We also appreciate the twelve evaluators who contributed to our human evaluation.

\bibliography{tacl2021}

\begin{thebibliography}{60}
\expandafter\ifx\csname natexlab\endcsname\relax\def\natexlab#1{#1}\fi

\bibitem[{Asai et~al.(2024)Asai, Wu, Wang, Sil, and Hajishirzi}]{ref17}
Akari Asai, Zeqiu Wu, Yizhong Wang, Avirup Sil, and Hannaneh Hajishirzi. 2024.
\newblock \href {https://openreview.net/forum?id=hSyW5go0v8} {Self-{RAG}:
  Learning to retrieve, generate, and critique through self-reflection}.
\newblock In \emph{The Twelfth International Conference on Learning
  Representations}.

\bibitem[{Bai et~al.(2023)Bai, Yan, Yang, Yang, Liang, Guo, and Li}]{ref30}
Jiaqi Bai, Zhao Yan, Ze~Yang, Jian Yang, Xinnian Liang, Hongcheng Guo, and
  Zhoujun Li. 2023.
\newblock \href {https://doi.org/10.1007/978-3-031-43415-0_31}
  {Knowprefix-tuning: A two-stage prefix-tuning framework
  for knowledge-grounded dialogue generation}.
\newblock In \emph{Machine Learning and Knowledge Discovery in Databases:
  Research Track}, pages 525--542, Cham. Springer Nature Switzerland.

\bibitem[{Barber and Agakov(2003)}]{ref42}
David Barber and Felix Agakov. 2003.
\newblock \href
  {https://proceedings.neurips.cc/paper_files/paper/2003/file/a6ea8471c120fe8cc35a2954c9b9c595-Paper.pdf}
  {The im algorithm: a variational approach to information maximization}.
\newblock In \emph{Advances in Neural Information Processing Systems},
  volume~16. MIT Press.

\bibitem[{Borgeaud et~al.(2022)Borgeaud, Mensch, Hoffmann, Cai, Rutherford,
  Millican, Van Den~Driessche, Lespiau, Damoc, Clark, De~Las~Casas, Guy,
  Menick, Ring, Hennigan, Huang, Maggiore, Jones, Cassirer, Brock, Paganini,
  Irving, Vinyals, Osindero, Simonyan, Rae, Elsen, and Sifre}]{ref10}
Sebastian Borgeaud, Arthur Mensch, Jordan Hoffmann, Trevor Cai, Eliza
  Rutherford, Katie Millican, George~Bm Van Den~Driessche, Jean-Baptiste
  Lespiau, Bogdan Damoc, Aidan Clark, Diego De~Las~Casas, Aurelia Guy, Jacob
  Menick, Roman Ring, Tom Hennigan, Saffron Huang, Loren Maggiore, Chris Jones,
  Albin Cassirer, Andy Brock, Michela Paganini, Geoffrey Irving, Oriol Vinyals,
  Simon Osindero, Karen Simonyan, Jack Rae, Erich Elsen, and Laurent Sifre.
  2022.
\newblock \href {https://proceedings.mlr.press/v162/borgeaud22a.html}
  {Improving language models by retrieving from trillions of tokens}.
\newblock In \emph{Proceedings of the 39th International Conference on Machine
  Learning}, volume 162 of \emph{Proceedings of Machine Learning Research},
  pages 2206--2240. PMLR.

\bibitem[{Brown et~al.(2020)Brown, Mann, Ryder, Subbiah, Kaplan, Dhariwal,
  Neelakantan, Shyam, Sastry, Askell, Agarwal, Herbert-Voss, Krueger, Henighan,
  Child, Ramesh, Ziegler, Wu, Winter, Hesse, Chen, Sigler, Litwin, Gray, Chess,
  Clark, Berner, McCandlish, Radford, Sutskever, and Amodei}]{ref1}
Tom Brown, Benjamin Mann, Nick Ryder, Melanie Subbiah, Jared~D Kaplan, Prafulla
  Dhariwal, Arvind Neelakantan, Pranav Shyam, Girish Sastry, Amanda Askell,
  Sandhini Agarwal, Ariel Herbert-Voss, Gretchen Krueger, Tom Henighan, Rewon
  Child, Aditya Ramesh, Daniel Ziegler, Jeffrey Wu, Clemens Winter, Chris
  Hesse, Mark Chen, Eric Sigler, Mateusz Litwin, Scott Gray, Benjamin Chess,
  Jack Clark, Christopher Berner, Sam McCandlish, Alec Radford, Ilya Sutskever,
  and Dario Amodei. 2020.
\newblock \href
  {https://proceedings.neurips.cc/paper_files/paper/2020/file/1457c0d6bfcb4967418bfb8ac142f64a-Paper.pdf}
  {Language models are few-shot learners}.
\newblock In \emph{Advances in Neural Information Processing Systems},
  volume~33, pages 1877--1901. Curran Associates, Inc.

\bibitem[{Chowdhery et~al.(2023)Chowdhery, Narang, Devlin, Bosma, Mishra,
  Roberts, Barham, Chung, Sutton, Gehrmann, Schuh, Shi, Tsvyashchenko, Maynez,
  Rao, Barnes, Tay, Shazeer, Prabhakaran, Reif, Du, Hutchinson, Pope, Bradbury,
  Austin, Isard, Gur-Ari, Yin, Duke, Levskaya, Ghemawat, Dev, Michalewski,
  Garcia, Misra, Robinson, Fedus, Zhou, Ippolito, Luan, Lim, Zoph, Spiridonov,
  Sepassi, Dohan, Agrawal, Omernick, Dai, Pillai, Pellat, Lewkowycz, Moreira,
  Child, Polozov, Lee, Zhou, Wang, Saeta, Diaz, Firat, Catasta, Wei,
  Meier-Hellstern, Eck, Dean, Petrov, and Fiedel}]{ref5}
Aakanksha Chowdhery, Sharan Narang, Jacob Devlin, Maarten Bosma, Gaurav Mishra,
  Adam Roberts, Paul Barham, Hyung~Won Chung, Charles Sutton, Sebastian
  Gehrmann, Parker Schuh, Kensen Shi, Sasha Tsvyashchenko, Joshua Maynez,
  Abhishek Rao, Parker Barnes, Yi~Tay, Noam Shazeer, Vinodkumar Prabhakaran,
  Emily Reif, Nan Du, Ben Hutchinson, Reiner Pope, James Bradbury, Jacob
  Austin, Michael Isard, Guy Gur-Ari, Pengcheng Yin, Toju Duke, Anselm
  Levskaya, Sanjay Ghemawat, Sunipa Dev, Henryk Michalewski, Xavier Garcia,
  Vedant Misra, Kevin Robinson, Liam Fedus, Denny Zhou, Daphne Ippolito, David
  Luan, Hyeontaek Lim, Barret Zoph, Alexander Spiridonov, Ryan Sepassi, David
  Dohan, Shivani Agrawal, Mark Omernick, Andrew~M. Dai,
  Thanumalayan~Sankaranarayana Pillai, Marie Pellat, Aitor Lewkowycz, Erica
  Moreira, Rewon Child, Oleksandr Polozov, Katherine Lee, Zongwei Zhou, Xuezhi
  Wang, Brennan Saeta, Mark Diaz, Orhan Firat, Michele Catasta, Jason Wei,
  Kathy Meier-Hellstern, Douglas Eck, Jeff Dean, Slav Petrov, and Noah Fiedel.
  2023.
\newblock \href {http://jmlr.org/papers/v24/22-1144.html} {Palm: Scaling
  language modeling with pathways}.
\newblock \emph{Journal of Machine Learning Research}, 24(240):1--113.

\bibitem[{Cohen(1960)}]{ref45}
Jacob Cohen. 1960.
\newblock \href {https://doi.org/10.1177/001316446002000104} {A coefficient of
  agreement for nominal scales}.
\newblock \emph{Educational and Psychological Measurement}, 20(1):37--46.

\bibitem[{Devlin et~al.(2019)Devlin, Chang, Lee, and Toutanova}]{ref41}
Jacob Devlin, Ming-Wei Chang, Kenton Lee, and Kristina Toutanova. 2019.
\newblock \href {https://doi.org/10.18653/v1/N19-1423} {{BERT}: Pre-training of
  deep bidirectional transformers for language understanding}.
\newblock In \emph{Proceedings of the 2019 Conference of the North {A}merican
  Chapter of the Association for Computational Linguistics: Human Language
  Technologies, Volume 1 (Long and Short Papers)}, pages 4171--4186,
  Minneapolis, Minnesota. Association for Computational Linguistics.

\bibitem[{Dinan et~al.(2019)Dinan, Roller, Shuster, Fan, Auli, and
  Weston}]{ref24}
Emily Dinan, Stephen Roller, Kurt Shuster, Angela Fan, Michael Auli, and Jason
  Weston. 2019.
\newblock \href {https://openreview.net/forum?id=r1l73iRqKm} {Wizard of
  wikipedia: Knowledge-powered conversational agents}.
\newblock In \emph{International Conference on Learning Representations}.

\bibitem[{Fan et~al.(2019)Fan, Jernite, Perez, Grangier, Weston, and
  Auli}]{ref53}
Angela Fan, Yacine Jernite, Ethan Perez, David Grangier, Jason Weston, and
  Michael Auli. 2019.
\newblock \href {https://doi.org/10.18653/v1/P19-1346} {{ELI}5: Long form
  question answering}.
\newblock In \emph{Proceedings of the 57th Annual Meeting of the Association
  for Computational Linguistics}, pages 3558--3567, Florence, Italy.
  Association for Computational Linguistics.

\bibitem[{Guu et~al.(2020)Guu, Lee, Tung, Pasupat, and Chang}]{ref9}
Kelvin Guu, Kenton Lee, Zora Tung, Panupong Pasupat, and Mingwei Chang. 2020.
\newblock \href {https://proceedings.mlr.press/v119/guu20a.html} {Retrieval
  augmented language model pre-training}.
\newblock In \emph{Proceedings of the 37th International Conference on Machine
  Learning}, volume 119 of \emph{Proceedings of Machine Learning Research},
  pages 3929--3938. PMLR.

\bibitem[{He et~al.(2022)He, Zhou, Ma, Berg-Kirkpatrick, and Neubig}]{ref40}
Junxian He, Chunting Zhou, Xuezhe Ma, Taylor Berg-Kirkpatrick, and Graham
  Neubig. 2022.
\newblock \href {https://openreview.net/forum?id=0RDcd5Axok} {Towards a unified
  view of parameter-efficient transfer learning}.
\newblock In \emph{International Conference on Learning Representations}.

\bibitem[{He et~al.(2023)He, Gao, and Chen}]{ref55}
Pengcheng He, Jianfeng Gao, and Weizhu Chen. 2023.
\newblock \href {https://openreview.net/forum?id=sE7-XhLxHA} {De{BERT}av3:
  Improving de{BERT}a using {ELECTRA}-style pre-training with
  gradient-disentangled embedding sharing}.
\newblock In \emph{The Eleventh International Conference on Learning
  Representations}.

\bibitem[{Hoffart et~al.(2011)Hoffart, Yosef, Bordino, F{\"u}rstenau, Pinkal,
  Spaniol, Taneva, Thater, and Weikum}]{ref48}
Johannes Hoffart, Mohamed~Amir Yosef, Ilaria Bordino, Hagen F{\"u}rstenau,
  Manfred Pinkal, Marc Spaniol, Bilyana Taneva, Stefan Thater, and Gerhard
  Weikum. 2011.
\newblock \href {https://aclanthology.org/D11-1072} {Robust disambiguation of
  named entities in text}.
\newblock In \emph{Proceedings of the 2011 Conference on Empirical Methods in
  Natural Language Processing}, pages 782--792, Edinburgh, Scotland, UK.
  Association for Computational Linguistics.

\bibitem[{Houlsby et~al.(2019)Houlsby, Giurgiu, Jastrzebski, Morrone,
  De~Laroussilhe, Gesmundo, Attariyan, and Gelly}]{ref31}
Neil Houlsby, Andrei Giurgiu, Stanislaw Jastrzebski, Bruna Morrone, Quentin
  De~Laroussilhe, Andrea Gesmundo, Mona Attariyan, and Sylvain Gelly. 2019.
\newblock \href {https://proceedings.mlr.press/v97/houlsby19a.html}
  {Parameter-efficient transfer learning for {NLP}}.
\newblock In \emph{Proceedings of the 36th International Conference on Machine
  Learning}, volume~97 of \emph{Proceedings of Machine Learning Research},
  pages 2790--2799. PMLR.

\bibitem[{Hu et~al.(2022)Hu, yelong shen, Wallis, Allen-Zhu, Li, Wang, Wang,
  and Chen}]{ref33}
Edward~J Hu, yelong shen, Phillip Wallis, Zeyuan Allen-Zhu, Yuanzhi Li, Shean
  Wang, Lu~Wang, and Weizhu Chen. 2022.
\newblock \href {https://openreview.net/forum?id=nZeVKeeFYf9} {Lo{RA}: Low-rank
  adaptation of large language models}.
\newblock In \emph{International Conference on Learning Representations}.

\bibitem[{Izacard et~al.(2022)Izacard, Caron, Hosseini, Riedel, Bojanowski,
  Joulin, and Grave}]{ref54}
Gautier Izacard, Mathilde Caron, Lucas Hosseini, Sebastian Riedel, Piotr
  Bojanowski, Armand Joulin, and Edouard Grave. 2022.
\newblock \href {https://openreview.net/forum?id=jKN1pXi7b0} {Unsupervised
  dense information retrieval with contrastive learning}.
\newblock \emph{Transactions on Machine Learning Research}.

\bibitem[{Izacard et~al.(2023)Izacard, Lewis, Lomeli, Hosseini, Petroni,
  Schick, Dwivedi-Yu, Joulin, Riedel, and Grave}]{ref13}
Gautier Izacard, Patrick Lewis, Maria Lomeli, Lucas Hosseini, Fabio Petroni,
  Timo Schick, Jane Dwivedi-Yu, Armand Joulin, Sebastian Riedel, and Edouard
  Grave. 2023.
\newblock \href {http://jmlr.org/papers/v24/23-0037.html} {Atlas: Few-shot
  learning with retrieval augmented language models}.
\newblock \emph{Journal of Machine Learning Research}, 24(251):1--43.

\bibitem[{Jiang et~al.(2023)Jiang, Sablayrolles, Mensch, Bamford, Chaplot,
  de~las Casas, Bressand, Lengyel, Lample, Saulnier, Lavaud, Lachaux, Stock,
  Scao, Lavril, Wang, Lacroix, and Sayed}]{jiang2023mistral7b}
Albert~Q. Jiang, Alexandre Sablayrolles, Arthur Mensch, Chris Bamford,
  Devendra~Singh Chaplot, Diego de~las Casas, Florian Bressand, Gianna Lengyel,
  Guillaume Lample, Lucile Saulnier, Lélio~Renard Lavaud, Marie-Anne Lachaux,
  Pierre Stock, Teven~Le Scao, Thibaut Lavril, Thomas Wang, Timothée Lacroix,
  and William~El Sayed. 2023.
\newblock \href {https://doi.org/10.48550/arXiv.2310.06825} {Mistral 7b}.
\newblock \emph{CoRR}, abs/2310.06825v1.

\bibitem[{Jin et~al.(2019)Jin, Dhingra, Liu, Cohen, and Lu}]{ref43}
Qiao Jin, Bhuwan Dhingra, Zhengping Liu, William Cohen, and Xinghua Lu. 2019.
\newblock \href {https://doi.org/10.18653/v1/D19-1259} {{P}ub{M}ed{QA}: A
  dataset for biomedical research question answering}.
\newblock In \emph{Proceedings of the 2019 Conference on Empirical Methods in
  Natural Language Processing and the 9th International Joint Conference on
  Natural Language Processing (EMNLP-IJCNLP)}, pages 2567--2577, Hong Kong,
  China. Association for Computational Linguistics.

\bibitem[{Joshi et~al.(2017)Joshi, Choi, Weld, and Zettlemoyer}]{ref52}
Mandar Joshi, Eunsol Choi, Daniel Weld, and Luke Zettlemoyer. 2017.
\newblock \href {https://doi.org/10.18653/v1/P17-1147} {{T}rivia{QA}: A large
  scale distantly supervised challenge dataset for reading comprehension}.
\newblock In \emph{Proceedings of the 55th Annual Meeting of the Association
  for Computational Linguistics (Volume 1: Long Papers)}, pages 1601--1611,
  Vancouver, Canada. Association for Computational Linguistics.

\bibitem[{Kandpal et~al.(2023)Kandpal, Deng, Roberts, Wallace, and
  Raffel}]{ref6}
Nikhil Kandpal, Haikang Deng, Adam Roberts, Eric Wallace, and Colin Raffel.
  2023.
\newblock \href {https://proceedings.mlr.press/v202/kandpal23a.html} {Large
  language models struggle to learn long-tail knowledge}.
\newblock In \emph{Proceedings of the 40th International Conference on Machine
  Learning}, volume 202 of \emph{Proceedings of Machine Learning Research},
  pages 15696--15707. PMLR.

\bibitem[{Karimi~Mahabadi et~al.(2021)Karimi~Mahabadi, Henderson, and
  Ruder}]{ref32}
Rabeeh Karimi~Mahabadi, James Henderson, and Sebastian Ruder. 2021.
\newblock \href
  {https://proceedings.neurips.cc/paper_files/paper/2021/file/081be9fdff07f3bc808f935906ef70c0-Paper.pdf}
  {Compacter: Efficient low-rank hypercomplex adapter layers}.
\newblock In \emph{Advances in Neural Information Processing Systems},
  volume~34, pages 1022--1035. Curran Associates, Inc.

\bibitem[{Kim et~al.(2020)Kim, Ahn, and Kim}]{ref27}
Byeongchang Kim, Jaewoo Ahn, and Gunhee Kim. 2020.
\newblock \href {https://openreview.net/forum?id=Hke0K1HKwr} {Sequential latent
  knowledge selection for knowledge-grounded dialogue}.
\newblock In \emph{International Conference on Learning Representations}.

\bibitem[{Kwiatkowski et~al.(2019)Kwiatkowski, Palomaki, Redfield, Collins,
  Parikh, Alberti, Epstein, Polosukhin, Devlin, Lee, Toutanova, Jones, Kelcey,
  Chang, Dai, Uszkoreit, Le, and Petrov}]{ref50}
Tom Kwiatkowski, Jennimaria Palomaki, Olivia Redfield, Michael Collins, Ankur
  Parikh, Chris Alberti, Danielle Epstein, Illia Polosukhin, Jacob Devlin,
  Kenton Lee, Kristina Toutanova, Llion Jones, Matthew Kelcey, Ming-Wei Chang,
  Andrew~M. Dai, Jakob Uszkoreit, Quoc Le, and Slav Petrov. 2019.
\newblock \href {https://doi.org/10.1162/tacl_a_00276} {Natural questions: A
  benchmark for question answering research}.
\newblock \emph{Transactions of the Association for Computational Linguistics},
  7:452--466.

\bibitem[{Lester et~al.(2021)Lester, Al-Rfou, and Constant}]{ref36}
Brian Lester, Rami Al-Rfou, and Noah Constant. 2021.
\newblock \href {https://doi.org/10.18653/v1/2021.emnlp-main.243} {The power of
  scale for parameter-efficient prompt tuning}.
\newblock In \emph{Proceedings of the 2021 Conference on Empirical Methods in
  Natural Language Processing}, pages 3045--3059, Online and Punta Cana,
  Dominican Republic. Association for Computational Linguistics.

\bibitem[{Levy et~al.(2017)Levy, Seo, Choi, and Zettlemoyer}]{ref49}
Omer Levy, Minjoon Seo, Eunsol Choi, and Luke Zettlemoyer. 2017.
\newblock \href {https://doi.org/10.18653/v1/K17-1034} {Zero-shot relation
  extraction via reading comprehension}.
\newblock In \emph{Proceedings of the 21st Conference on Computational Natural
  Language Learning ({C}o{NLL} 2017)}, pages 333--342, Vancouver, Canada.
  Association for Computational Linguistics.

\bibitem[{Lewis et~al.(2020{\natexlab{a}})Lewis, Liu, Goyal, Ghazvininejad,
  Mohamed, Levy, Stoyanov, and Zettlemoyer}]{ref25}
Mike Lewis, Yinhan Liu, Naman Goyal, Marjan Ghazvininejad, Abdelrahman Mohamed,
  Omer Levy, Veselin Stoyanov, and Luke Zettlemoyer. 2020{\natexlab{a}}.
\newblock \href {https://doi.org/10.18653/v1/2020.acl-main.703} {{BART}:
  Denoising sequence-to-sequence pre-training for natural language generation,
  translation, and comprehension}.
\newblock In \emph{Proceedings of the 58th Annual Meeting of the Association
  for Computational Linguistics}, pages 7871--7880, Online. Association for
  Computational Linguistics.

\bibitem[{Lewis et~al.(2020{\natexlab{b}})Lewis, Perez, Piktus, Petroni,
  Karpukhin, Goyal, K\"{u}ttler, Lewis, Yih, Rockt\"{a}schel, Riedel, and
  Kiela}]{ref8}
Patrick Lewis, Ethan Perez, Aleksandra Piktus, Fabio Petroni, Vladimir
  Karpukhin, Naman Goyal, Heinrich K\"{u}ttler, Mike Lewis, Wen-tau Yih, Tim
  Rockt\"{a}schel, Sebastian Riedel, and Douwe Kiela. 2020{\natexlab{b}}.
\newblock \href
  {https://proceedings.neurips.cc/paper_files/paper/2020/file/6b493230205f780e1bc26945df7481e5-Paper.pdf}
  {Retrieval-augmented generation for knowledge-intensive nlp tasks}.
\newblock In \emph{Advances in Neural Information Processing Systems},
  volume~33, pages 9459--9474. Curran Associates, Inc.

\bibitem[{Li et~al.(2023)Li, Li, Savarese, and Hoi}]{ref22}
Junnan Li, Dongxu Li, Silvio Savarese, and Steven Hoi. 2023.
\newblock \href {https://proceedings.mlr.press/v202/li23q.html} {{BLIP}-2:
  Bootstrapping language-image pre-training with frozen image encoders and
  large language models}.
\newblock In \emph{Proceedings of the 40th International Conference on Machine
  Learning}, volume 202 of \emph{Proceedings of Machine Learning Research},
  pages 19730--19742. PMLR.

\bibitem[{Li and Liang(2021)}]{ref37}
Xiang~Lisa Li and Percy Liang. 2021.
\newblock \href {https://doi.org/10.18653/v1/2021.acl-long.353} {Prefix-tuning:
  Optimizing continuous prompts for generation}.
\newblock In \emph{Proceedings of the 59th Annual Meeting of the Association
  for Computational Linguistics and the 11th International Joint Conference on
  Natural Language Processing (Volume 1: Long Papers)}, pages 4582--4597,
  Online. Association for Computational Linguistics.

\bibitem[{Lin et~al.(2024)Lin, Chen, Chen, Shi, Lomeli, James, Rodriguez, Kahn,
  Szilvasy, Lewis, Zettlemoyer, and tau Yih}]{ref16}
Xi~Victoria Lin, Xilun Chen, Mingda Chen, Weijia Shi, Maria Lomeli, Richard
  James, Pedro Rodriguez, Jacob Kahn, Gergely Szilvasy, Mike Lewis, Luke
  Zettlemoyer, and Wen tau Yih. 2024.
\newblock \href {https://openreview.net/forum?id=22OTbutug9} {{RA}-{DIT}:
  Retrieval-augmented dual instruction tuning}.
\newblock In \emph{The Twelfth International Conference on Learning
  Representations}.

\bibitem[{Liu et~al.(2016)Liu, Lowe, Serban, Noseworthy, Charlin, and
  Pineau}]{liu-etal-2016-evaluate}
Chia-Wei Liu, Ryan Lowe, Iulian Serban, Mike Noseworthy, Laurent Charlin, and
  Joelle Pineau. 2016.
\newblock \href {https://doi.org/10.18653/v1/D16-1230} {How {NOT} to evaluate
  your dialogue system: An empirical study of unsupervised evaluation metrics
  for dialogue response generation}.
\newblock In \emph{Proceedings of the 2016 Conference on Empirical Methods in
  Natural Language Processing}, pages 2122--2132, Austin, Texas. Association
  for Computational Linguistics.

\bibitem[{Liu et~al.(2021)Liu, Zhao, Li, Ren, Zhang, and Yin}]{ref29}
Shilei Liu, Xiaofeng Zhao, Bochao Li, Feiliang Ren, Longhui Zhang, and Shujuan
  Yin. 2021.
\newblock \href {https://doi.org/10.18653/v1/2021.emnlp-main.173} {{A}
  {T}hree-{S}tage {L}earning {F}ramework for {L}ow-{R}esource
  {K}nowledge-{G}rounded {D}ialogue {G}eneration}.
\newblock In \emph{Proceedings of the 2021 Conference on Empirical Methods in
  Natural Language Processing}, pages 2262--2272, Online and Punta Cana,
  Dominican Republic. Association for Computational Linguistics.

\bibitem[{Liu et~al.(2022)Liu, Ji, Fu, Tam, Du, Yang, and Tang}]{ref39}
Xiao Liu, Kaixuan Ji, Yicheng Fu, Weng Tam, Zhengxiao Du, Zhilin Yang, and Jie
  Tang. 2022.
\newblock \href {https://doi.org/10.18653/v1/2022.acl-short.8} {{P}-tuning:
  Prompt tuning can be comparable to fine-tuning across scales and tasks}.
\newblock In \emph{Proceedings of the 60th Annual Meeting of the Association
  for Computational Linguistics (Volume 2: Short Papers)}, pages 61--68,
  Dublin, Ireland. Association for Computational Linguistics.

\bibitem[{Liu et~al.(2024)Liu, Zheng, Du, Ding, Qian, Yang, and Tang}]{ref38}
Xiao Liu, Yanan Zheng, Zhengxiao Du, Ming Ding, Yujie Qian, Zhilin Yang, and
  Jie Tang. 2024.
\newblock \href {https://doi.org/https://doi.org/10.1016/j.aiopen.2023.08.012}
  {Gpt understands, too}.
\newblock \emph{AI Open}, 5:208--215.

\bibitem[{Meng et~al.(2021)Meng, Ren, Chen, Ren, Xi, and Rijke}]{ref18}
Chuan Meng, Pengjie Ren, Zhumin Chen, Zhaochun Ren, Tengxiao Xi, and Maarten~de
  Rijke. 2021.
\newblock \href {https://doi.org/10.1145/3404835.3462824} {Initiative-aware
  self-supervised learning for knowledge-grounded conversations}.
\newblock In \emph{Proceedings of the 44th International ACM SIGIR Conference
  on Research and Development in Information Retrieval}, SIGIR '21, pages
  522--532, New York, NY, USA. Association for Computing Machinery.

\bibitem[{Novikova et~al.(2017)Novikova, Du{\v{s}}ek, Cercas~Curry, and
  Rieser}]{novikova-etal-2017-need}
Jekaterina Novikova, Ond{\v{r}}ej Du{\v{s}}ek, Amanda Cercas~Curry, and Verena
  Rieser. 2017.
\newblock \href {https://doi.org/10.18653/v1/D17-1238} {Why we need new
  evaluation metrics for {NLG}}.
\newblock In \emph{Proceedings of the 2017 Conference on Empirical Methods in
  Natural Language Processing}, pages 2241--2252, Copenhagen, Denmark.
  Association for Computational Linguistics.

\bibitem[{OpenAI(2022)}]{ref2}
OpenAI.
\newblock \href {https://openai.com/blog/chatgpt/} {Introducing {ChatGPT}}
  [online]. 2022.

\bibitem[{OpenAI(2023)}]{ref3}
OpenAI. 2023.
\newblock \href {https://doi.org/10.48550/ARXIV.2303.08774} {{GPT-4} technical
  report}.
\newblock \emph{CoRR}, abs/2303.08774v3.

\bibitem[{Petroni et~al.(2021)Petroni, Piktus, Fan, Lewis, Yazdani, De~Cao,
  Thorne, Jernite, Karpukhin, Maillard, Plachouras, Rockt{\"a}schel, and
  Riedel}]{ref46}
Fabio Petroni, Aleksandra Piktus, Angela Fan, Patrick Lewis, Majid Yazdani,
  Nicola De~Cao, James Thorne, Yacine Jernite, Vladimir Karpukhin, Jean
  Maillard, Vassilis Plachouras, Tim Rockt{\"a}schel, and Sebastian Riedel.
  2021.
\newblock \href {https://doi.org/10.18653/v1/2021.naacl-main.200} {{KILT}: a
  benchmark for knowledge intensive language tasks}.
\newblock In \emph{Proceedings of the 2021 Conference of the North American
  Chapter of the Association for Computational Linguistics: Human Language
  Technologies}, pages 2523--2544, Online. Association for Computational
  Linguistics.

\bibitem[{Raffel et~al.(2020)Raffel, Shazeer, Roberts, Lee, Narang, Matena,
  Zhou, Li, and Liu}]{ref26}
Colin Raffel, Noam Shazeer, Adam Roberts, Katherine Lee, Sharan Narang, Michael
  Matena, Yanqi Zhou, Wei Li, and Peter~J. Liu. 2020.
\newblock \href {http://jmlr.org/papers/v21/20-074.html} {Exploring the limits
  of transfer learning with a unified text-to-text transformer}.
\newblock \emph{Journal of Machine Learning Research}, 21(140):1--67.

\bibitem[{Ram et~al.(2023)Ram, Levine, Dalmedigos, Muhlgay, Shashua,
  Leyton-Brown, and Shoham}]{ref12}
Ori Ram, Yoav Levine, Itay Dalmedigos, Dor Muhlgay, Amnon Shashua, Kevin
  Leyton-Brown, and Yoav Shoham. 2023.
\newblock \href {https://doi.org/10.1162/tacl_a_00605} {{In-Context
  Retrieval-Augmented Language Models}}.
\newblock \emph{Transactions of the Association for Computational Linguistics},
  11:1316--1331.

\bibitem[{Sarthi et~al.(2024)Sarthi, Abdullah, Tuli, Khanna, Goldie, and
  Manning}]{ref15}
Parth Sarthi, Salman Abdullah, Aditi Tuli, Shubh Khanna, Anna Goldie, and
  Christopher~D Manning. 2024.
\newblock \href {https://openreview.net/forum?id=GN921JHCRw} {{RAPTOR}:
  Recursive abstractive processing for tree-organized retrieval}.
\newblock In \emph{The Twelfth International Conference on Learning
  Representations}.

\bibitem[{Shen et~al.(2022)Shen, Perez-Rosas, Welch, Poria, and
  Mihalcea}]{ref28}
Siqi Shen, Veronica Perez-Rosas, Charles Welch, Soujanya Poria, and Rada
  Mihalcea. 2022.
\newblock \href {https://doi.org/10.18653/v1/2022.acl-long.221} {Knowledge
  enhanced reflection generation for counseling dialogues}.
\newblock In \emph{Proceedings of the 60th Annual Meeting of the Association
  for Computational Linguistics (Volume 1: Long Papers)}, pages 3096--3107,
  Dublin, Ireland. Association for Computational Linguistics.

\bibitem[{Shi et~al.(2024)Shi, Min, Yasunaga, Seo, James, Lewis, Zettlemoyer,
  and Yih}]{ref21}
Weijia Shi, Sewon Min, Michihiro Yasunaga, Minjoon Seo, Richard James, Mike
  Lewis, Luke Zettlemoyer, and Wen-tau Yih. 2024.
\newblock \href {https://doi.org/10.18653/v1/2024.naacl-long.463} {{REPLUG}:
  Retrieval-augmented black-box language models}.
\newblock In \emph{Proceedings of the 2024 Conference of the North American
  Chapter of the Association for Computational Linguistics: Human Language
  Technologies (Volume 1: Long Papers)}, pages 8371--8384, Mexico City, Mexico.
  Association for Computational Linguistics.

\bibitem[{Siriwardhana et~al.(2023)Siriwardhana, Weerasekera, Wen,
  Kaluarachchi, Rana, and Nanayakkara}]{ref11}
Shamane Siriwardhana, Rivindu Weerasekera, Elliott Wen, Tharindu Kaluarachchi,
  Rajib Rana, and Suranga Nanayakkara. 2023.
\newblock \href {https://doi.org/10.1162/tacl_a_00530} {Improving the domain
  adaptation of retrieval augmented generation ({RAG}) models for open domain
  question answering}.
\newblock \emph{Transactions of the Association for Computational Linguistics},
  11:1--17.

\bibitem[{Thorne et~al.(2018)Thorne, Vlachos, Christodoulopoulos, and
  Mittal}]{ref47}
James Thorne, Andreas Vlachos, Christos Christodoulopoulos, and Arpit Mittal.
  2018.
\newblock \href {https://doi.org/10.18653/v1/N18-1074} {{FEVER}: a large-scale
  dataset for fact extraction and {VER}ification}.
\newblock In \emph{Proceedings of the 2018 Conference of the North {A}merican
  Chapter of the Association for Computational Linguistics: Human Language
  Technologies, Volume 1 (Long Papers)}, pages 809--819, New Orleans,
  Louisiana. Association for Computational Linguistics.

\bibitem[{Touvron et~al.(2023)Touvron, Lavril, Izacard, Martinet, Lachaux,
  Lacroix, Rozi{\`{e}}re, Goyal, Hambro, Azhar, Rodriguez, Joulin, Grave, and
  Lample}]{ref4}
Hugo Touvron, Thibaut Lavril, Gautier Izacard, Xavier Martinet, Marie{-}Anne
  Lachaux, Timoth{\'{e}}e Lacroix, Baptiste Rozi{\`{e}}re, Naman Goyal, Eric
  Hambro, Faisal Azhar, Aur{\'{e}}lien Rodriguez, Armand Joulin, Edouard Grave,
  and Guillaume Lample. 2023.
\newblock \href {https://doi.org/10.48550/ARXIV.2302.13971} {Llama: Open and
  efficient foundation language models}.
\newblock \emph{CoRR}, abs/2302.13971v1.

\bibitem[{Xu et~al.(2022)Xu, Ishii, Cahyawijaya, Liu, Winata, Madotto, Su, and
  Fung}]{xu-etal-2022-retrieval}
Yan Xu, Etsuko Ishii, Samuel Cahyawijaya, Zihan Liu, Genta~Indra Winata, Andrea
  Madotto, Dan Su, and Pascale Fung. 2022.
\newblock \href {https://doi.org/10.18653/v1/2022.dialdoc-1.10} {Retrieval-free
  knowledge-grounded dialogue response generation with adapters}.
\newblock In \emph{Proceedings of the Second DialDoc Workshop on
  Document-grounded Dialogue and Conversational Question Answering}, pages
  93--107, Dublin, Ireland. Association for Computational Linguistics.

\bibitem[{Xu et~al.(2023)Xu, Kong, Xu, Ji, Pang, Fung, and Wu}]{ref20}
Yan Xu, Deqian Kong, Dehong Xu, Ziwei Ji, Bo~Pang, Pascale Fung, and Ying~Nian
  Wu. 2023.
\newblock \href {https://proceedings.mlr.press/v202/xu23j.html} {Diverse and
  faithful knowledge-grounded dialogue generation via sequential posterior
  inference}.
\newblock In \emph{Proceedings of the 40th International Conference on Machine
  Learning}, volume 202 of \emph{Proceedings of Machine Learning Research},
  pages 38518--38534. PMLR.

\bibitem[{Yang et~al.(2025)Yang, Yang, Zhang, Hui, Zheng, Yu, Li, Liu, Huang,
  Wei, Lin, Yang, Tu, Zhang, Yang, Yang, Zhou, Lin, Dang, Lu, Bao, Yang, Yu,
  Li, Xue, Zhang, Zhu, Men, Lin, Li, Tang, Xia, Ren, Ren, Fan, Su, Zhang, Wan,
  Liu, Cui, Zhang, and Qiu}]{yang2024qwen2}
An~Yang, Baosong Yang, Beichen Zhang, Binyuan Hui, Bo~Zheng, Bowen Yu,
  Chengyuan Li, Dayiheng Liu, Fei Huang, Haoran Wei, Huan Lin, Jian Yang,
  Jianhong Tu, Jianwei Zhang, Jianxin Yang, Jiaxi Yang, Jingren Zhou, Junyang
  Lin, Kai Dang, Keming Lu, Keqin Bao, Kexin Yang, Le~Yu, Mei Li, Mingfeng Xue,
  Pei Zhang, Qin Zhu, Rui Men, Runji Lin, Tianhao Li, Tianyi Tang, Tingyu Xia,
  Xingzhang Ren, Xuancheng Ren, Yang Fan, Yang Su, Yichang Zhang, Yu~Wan,
  Yuqiong Liu, Zeyu Cui, Zhenru Zhang, and Zihan Qiu. 2025.
\newblock \href {https://doi.org/10.48550/arXiv.2412.15115} {Qwen2.5 technical
  report}.
\newblock \emph{CoRR}, abs/2412.15115v2.

\bibitem[{Yang et~al.(2018)Yang, Qi, Zhang, Bengio, Cohen, Salakhutdinov, and
  Manning}]{ref51}
Zhilin Yang, Peng Qi, Saizheng Zhang, Yoshua Bengio, William Cohen, Ruslan
  Salakhutdinov, and Christopher~D. Manning. 2018.
\newblock \href {https://doi.org/10.18653/v1/D18-1259} {{H}otpot{QA}: A dataset
  for diverse, explainable multi-hop question answering}.
\newblock In \emph{Proceedings of the 2018 Conference on Empirical Methods in
  Natural Language Processing}, pages 2369--2380, Brussels, Belgium.
  Association for Computational Linguistics.

\bibitem[{Yu et~al.(2023)Yu, Iter, Wang, Xu, Ju, Sanyal, Zhu, Zeng, and
  Jiang}]{ref14}
Wenhao Yu, Dan Iter, Shuohang Wang, Yichong Xu, Mingxuan Ju, Soumya Sanyal,
  Chenguang Zhu, Michael Zeng, and Meng Jiang. 2023.
\newblock \href {https://openreview.net/forum?id=fB0hRu9GZUS} {Generate rather
  than retrieve: Large language models are strong context generators}.
\newblock In \emph{The Eleventh International Conference on Learning
  Representations}.

\bibitem[{Zhang et~al.(2025)Zhang, Ma, Ding, Wang, Xu, and Lin}]{ref23}
Bo~Zhang, Hui Ma, Jian Ding, Jian Wang, Bo~Xu, and Hongfei Lin. 2025.
\newblock \href {https://doi.org/https://doi.org/10.1016/j.inffus.2025.102985}
  {Distilling implicit multimodal knowledge into large language models for
  zero-resource dialogue generation}.
\newblock \emph{Information Fusion}, 118:102985.

\bibitem[{Zhang et~al.(2022)Zhang, Wang, Lin, Ma, and Xu}]{ref19}
Bo~Zhang, Jian Wang, Hongfei Lin, Hui Ma, and Bo~Xu. 2022.
\newblock \href {https://doi.org/10.1109/TASLP.2022.3161151} {Exploiting
  pairwise mutual information for knowledge-grounded dialogue}.
\newblock \emph{IEEE/ACM Transactions on Audio, Speech, and Language
  Processing}, 30:2231--2240.

\bibitem[{Zhang et~al.(2023{\natexlab{a}})Zhang, Chen, Bukharin, He, Cheng,
  Chen, and Zhao}]{ref34}
Qingru Zhang, Minshuo Chen, Alexander Bukharin, Pengcheng He, Yu~Cheng, Weizhu
  Chen, and Tuo Zhao. 2023{\natexlab{a}}.
\newblock \href {https://openreview.net/forum?id=lq62uWRJjiY} {Adaptive budget
  allocation for parameter-efficient fine-tuning}.
\newblock In \emph{The Eleventh International Conference on Learning
  Representations}.

\bibitem[{Zhang et~al.(2024)Zhang, Han, Liu, Zhou, Lu, Qiao, Li, and
  Gao}]{ref35}
Renrui Zhang, Jiaming Han, Chris Liu, Aojun Zhou, Pan Lu, Yu~Qiao, Hongsheng
  Li, and Peng Gao. 2024.
\newblock \href {https://openreview.net/forum?id=d4UiXAHN2W}
  {{LL}a{MA}-adapter: Efficient fine-tuning of large language models with
  zero-initialized attention}.
\newblock In \emph{The Twelfth International Conference on Learning
  Representations}.

\bibitem[{Zhang et~al.(2023{\natexlab{b}})Zhang, Fang, Chen, Namazi-Rad, and
  Wang}]{ref7}
Zihan Zhang, Meng Fang, Ling Chen, Mohammad-Reza Namazi-Rad, and Jun Wang.
  2023{\natexlab{b}}.
\newblock \href {https://doi.org/10.18653/v1/2023.emnlp-main.516} {How do large
  language models capture the ever-changing world knowledge? a review of recent
  advances}.
\newblock In \emph{Proceedings of the 2023 Conference on Empirical Methods in
  Natural Language Processing}, pages 8289--8311, Singapore. Association for
  Computational Linguistics.

\bibitem[{Zheng et~al.(2023)Zheng, Chiang, Sheng, Zhuang, Wu, Zhuang, Lin, Li,
  Li, Xing, Zhang, Gonzalez, and Stoica}]{ref44}
Lianmin Zheng, Wei-Lin Chiang, Ying Sheng, Siyuan Zhuang, Zhanghao Wu, Yonghao
  Zhuang, Zi~Lin, Zhuohan Li, Dacheng Li, Eric Xing, Hao Zhang, Joseph~E
  Gonzalez, and Ion Stoica. 2023.
\newblock \href
  {https://proceedings.neurips.cc/paper_files/paper/2023/file/91f18a1287b398d378ef22505bf41832-Paper-Datasets_and_Benchmarks.pdf}
  {Judging llm-as-a-judge with mt-bench and chatbot arena}.
\newblock In \emph{Advances in Neural Information Processing Systems},
  volume~36, pages 46595--46623. Curran Associates, Inc.

\end{thebibliography}
\bibliographystyle{acl_natbib}

\onecolumn

\appendix

\section{KEDiT Training and Inference}
\label{appendix:algorithms}

Algorithms~\ref{alg:kedit_training} and~\ref{alg:kedit_inference} provide a step-by-step breakdown of our training and inference processes.

\begin{algorithm*}[ht!]
\small
\caption{KEDiT Training}
\label{alg:kedit_training}
\begin{algorithmic}[1]
    \State \textbf{Input:} Knowledge dataset $\mathcal{D}_k$, Knowledge-grounded dialogue dataset $\mathcal{D}_d$, Pre-trained BERT, Pre-trained LLM
    \Statex \textbf{Phase 1: Knowledge Compression Training}
    \State Initialize knowledge bottleneck with Q-Former and frozen BERT \Comment{Denoted as $\phi$}
    \State Freeze parameters of the pre-trained LLM \Comment{Denoted as $\psi$}
    \For{each $K \in \mathcal{D}_k$}
        \State Encode $K$ using the knowledge bottleneck to obtain $Z$ based on Eq.~\eqref{eq:z} \Comment{$p_\phi(Z|K)$}
        \State Reconstruct $K$ from $Z$ using the LLM to obtain $\hat{K}$ \Comment{$q_\psi(K|Z)$}
        \State Reconstruct $Z$ from $K$ using the LLM to obtain $\hat{Z}$ \Comment{$q_\psi(Z|K)$}
        \State Compute $\mathcal{L}_{\text{recon}}$ and $\mathcal{L}_{\text{align}}$ as defined in Eq.~\eqref{eq:recon} and Eq.~\eqref{eq:align}
        \State Update $\phi$ by minimizing $\mathcal{L}_{\text{kc}}$ as defined in Eq.~\eqref{eq:kc}
    \EndFor
    \Statex \textbf{Phase 2: Knowledge Integration Training}
    \State Integrate KA-Adapter with the frozen LLM \Comment{Denoted as $\theta$}
	\State Freeze $\phi$ except for $\mathbf{W}_z$
    \For{each $(C, K, R) \in \mathcal{D}_d$}
        \State Encode $K$ using the tuned knowledge bottleneck to obtain $Z$ \Comment{$p_\phi(Z|K)$}
        \State Concatenate $C$ and $Z$ to form the input
        \State Compute the likelihood of the response $R$ based on Eq.~\eqref{eq:ka-attn}, Eq.~\eqref{eq:ka-ffn} and Eq.~\eqref{eq:log} \Comment{$p_\theta(R|C, Z)$}
        \State Update $\theta$ and $\mathbf{W}_z$ by minimizing $\mathcal{L}_{\text{gen}}$ as defined in Eq.~\eqref{eq:gen}      
    \EndFor
    \State \textbf{Output:} Trained knowledge bottleneck $\phi$ and LLM integrated with KA-Adapter $\theta$
\end{algorithmic}
\end{algorithm*}

\begin{algorithm*}[ht!]
\small
\caption{KEDiT Inference}
\label{alg:kedit_inference}
\begin{algorithmic}[1]
    \State \textbf{Input:} Dialogue context $C$, Retrieved knowledge $K$, Trained knowledge bottleneck $\phi$, Trained LLM with KA-Adapter $\theta$
    \State Encode $K$ using the knowledge bottleneck module to obtain $Z$ \Comment{$p_\phi(Z|K)$}
	\State Initialize response $R_0 \gets \emptyset$, $t \gets 1$
    \While{not end-of-sequence}
        \State Concatenate $C$, $Z$ and $R_{<t}$ to form the input
        \State Compute the probability of the next token $R_t$ based on Eq.~\eqref{eq:ka-attn}, Eq.~\eqref{eq:ka-ffn} and Eq.~\eqref{eq:log} \Comment{$p_\theta(R_t|C, Z, R_{<t})$} 
        \State Sample or select $R_t$ based on $p_\theta(R_t|R_{<t}, C, Z)$
        \State Append $R_t$ to the response: $R \gets R \oplus R_t$
        \State Update $t \gets t + 1$
    \EndWhile
    \State \textbf{Output:} Generated response $R$
\end{algorithmic}
\end{algorithm*}


\section{Prompt Design and Iterative Validation Process for the PubMed-Dialog Dataset}
\label{appendix:prompt_example}

\paragraph{Iterative Validation Process}
In this work, we employ an iterative validation process to ensure the quality and faithfulness of the PubMed-Dialog dataset. This process involves the following steps:

\begin{enumerate}
    \item \textbf{First Round (Initial Evaluation)}: Each dialogue is evaluated on three key criteria: \textit{Source Consistency}, \textit{Internal Consistency}, and \textit{Factual Accuracy}. Scores ranging from 1 to 5 are assigned for each criterion, where 1 indicates poor quality and 5 represents high quality.
    \begin{itemize}
        \item \textit{Source Consistency}: Does the dialogue accurately and faithfully represent the information from the abstract?
        \item \textit{Internal Consistency}: Is the dialogue coherent and logically consistent within itself?
        \item \textit{Factual Accuracy}: Does the dialogue contain accurate medical information that is consistent with established knowledge?
    \end{itemize}
 
    \item \textbf{Regeneration}: Dialogues that score below a threshold (i.e., scores $<5$) are flagged for regeneration. These flagged dialogues are revised using GPT-4o, with a focus on eliminating hallucinations and ensuring a closer alignment with the source content, while adhering to the evaluation criteria.
    
    \item \textbf{Re-evaluation and Iterative Refinement}: The regenerated dialogues are evaluated again using the same criteria. If they still do not meet the required standards, they are flagged for further regeneration and re-evaluated until they meet the desired quality threshold.
\end{enumerate}

Table~\ref{tab:valid} summarizes the results of the three rounds of evaluation and regeneration for the PubMed-Dialog dataset. The iterative process has significantly improved dialogue quality across all criteria, particularly addressing initial deficiencies in \textit{Source Consistency}. While \textit{Internal Consistency} required minimal corrections and \textit{Factual Accuracy} improved substantially, minor alignment issues with biomedical knowledge persist, reflecting the complexity of the domain. Although the dataset does not achieve absolute perfection, it provides a robust benchmark for knowledge-grounded dialogue generation in specialized domains.

\begin{table*}[ht]
	\footnotesize
	\centering
	\textsc{\resizebox{\textwidth}{!}{
	\begin{tabular}{M{0.1\textwidth}M{0.16\textwidth}M{0.15\textwidth}M{0.15\textwidth}M{0.14\textwidth}M{0.15\textwidth}}
		\toprule
		Round & Source Consistency ($\leq 3/4/5$) & Internal Consistency ($\leq 3/4/5$) & Factual Accuracy ($\leq 3/4/5$) & Total Dialogues Evaluated & Percentage of High-Quality Dialogues \\
		\midrule
		Round 1 & 57 / 1,961 / 8,912 & 1 / 40 / 10,889 & 19 / 686 / 10,225 & 10,930 & 8,884 (81.28\%) \\
		Round 2 & 1 / 712 / 1,333 & 0 / 3 / 2,043 & 0 / 148 / 1,898  & 2,046 & 1,314 (64.22\%) \\
		Round 3 & 0 / 399 / 333 & 0 / 3 /729 & 0 / 70 / 662 & 732 & 321 (43.85\%)  \\
		\bottomrule
	\end{tabular}
	}}
	\caption{Results of the iterative evaluation and regeneration process, showing the number of dialogues scoring $\leq$ 3, 4, and 5 in each of the three evaluation criteria across all three validation rounds.}
	\label{tab:valid}
\end{table*}

\begin{figure*}[ht]
	\centering
	\includegraphics[width=\textwidth]{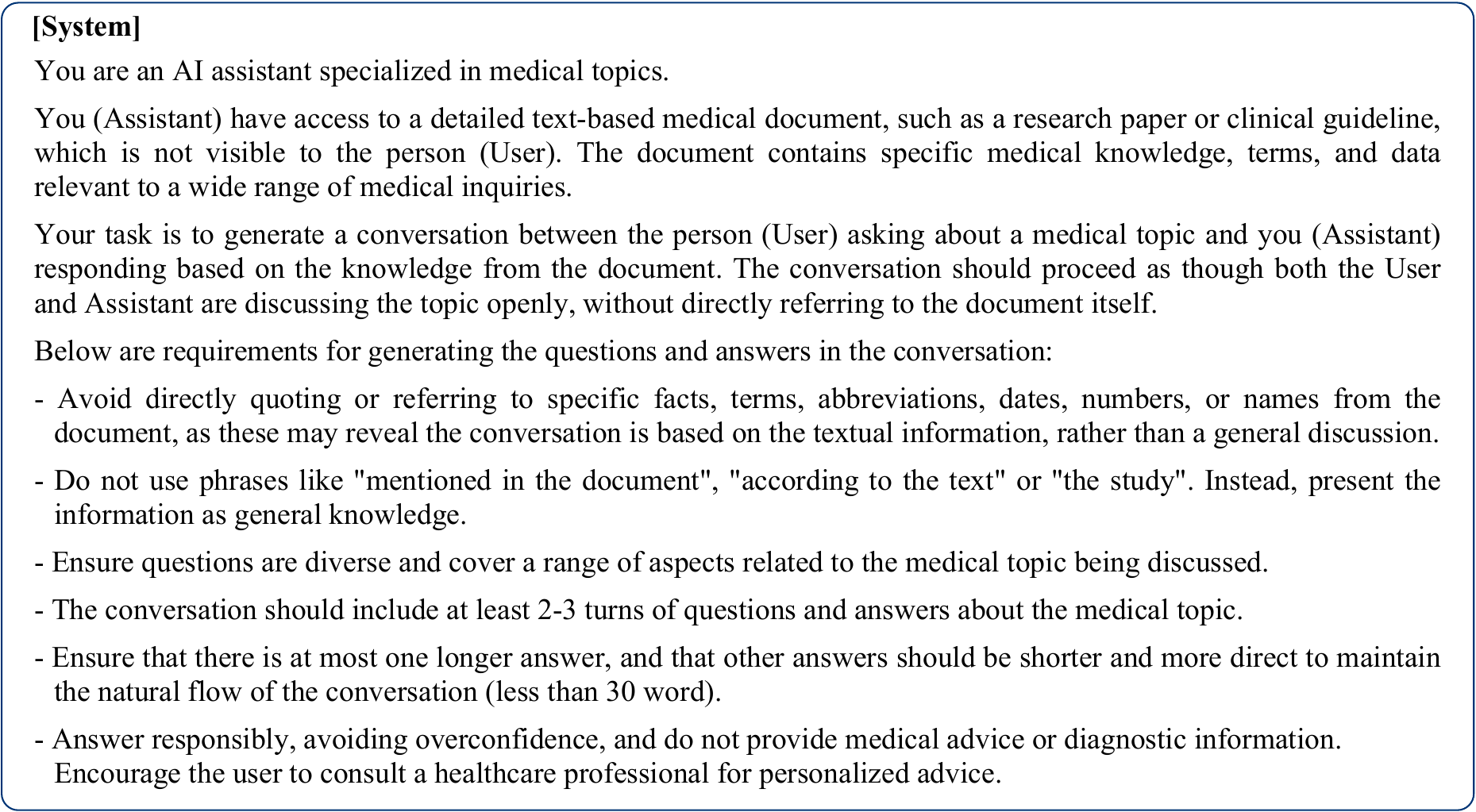}
	\caption{Example of the system prompt for generating PubMed-Dialog datasets.}
	\label{fig:prompt}
\end{figure*}

\paragraph{Prompt Design}
Figure~\ref{fig:prompt} shows the system prompt used to generate the PubMed-Dialog dataset.

\newpage

\section{Prompt Templates for LLM-Based Evaluation}
\label{appendix:prompt_template}

\begin{figure*}[ht!]
	\centering
	\includegraphics[width=\textwidth]{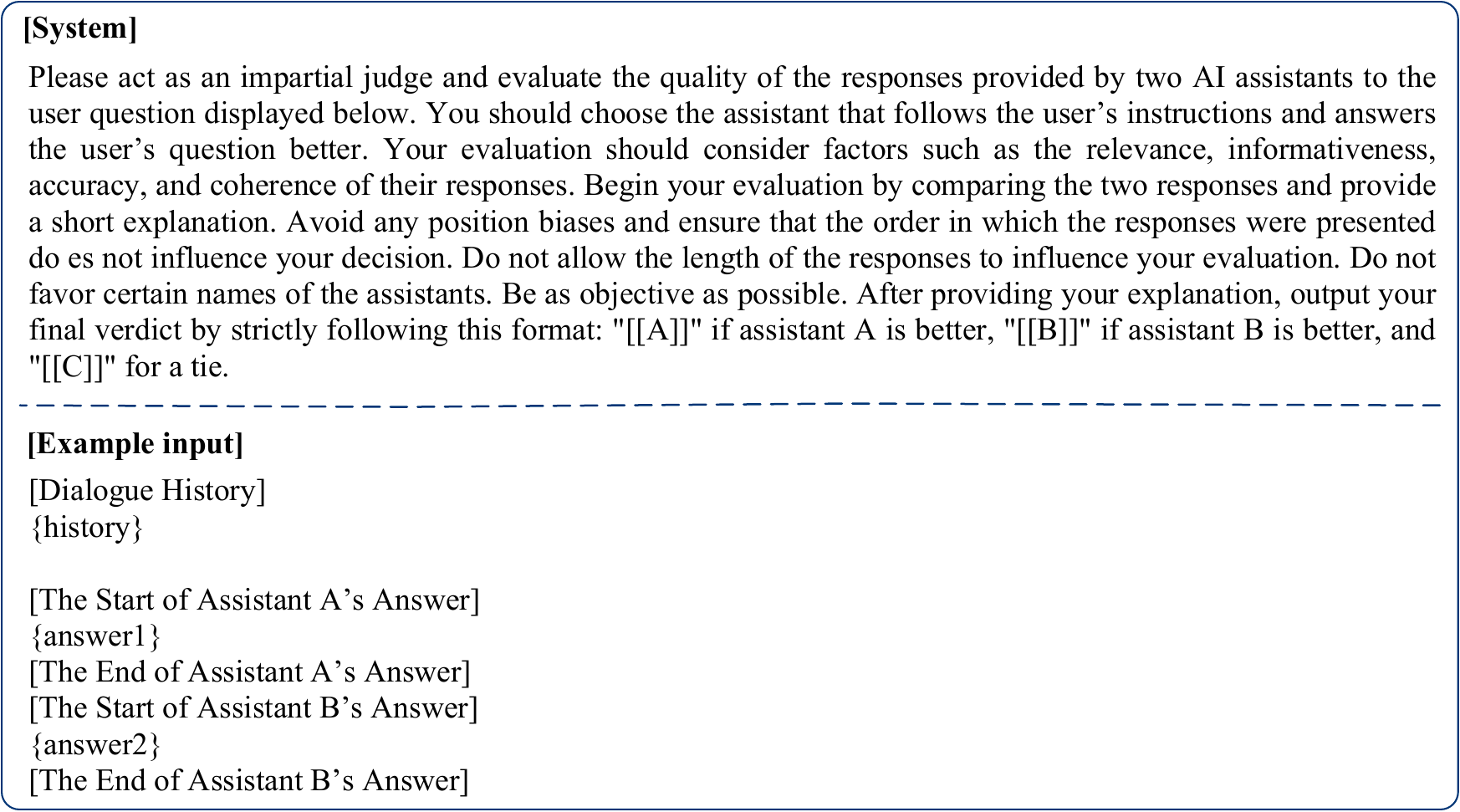}
	\caption{Example of the prompt used for pairwise comparison.}
	\label{fig:prompt2}
\end{figure*}

\begin{figure*}[h!]
	\centering
	\includegraphics[width=\textwidth]{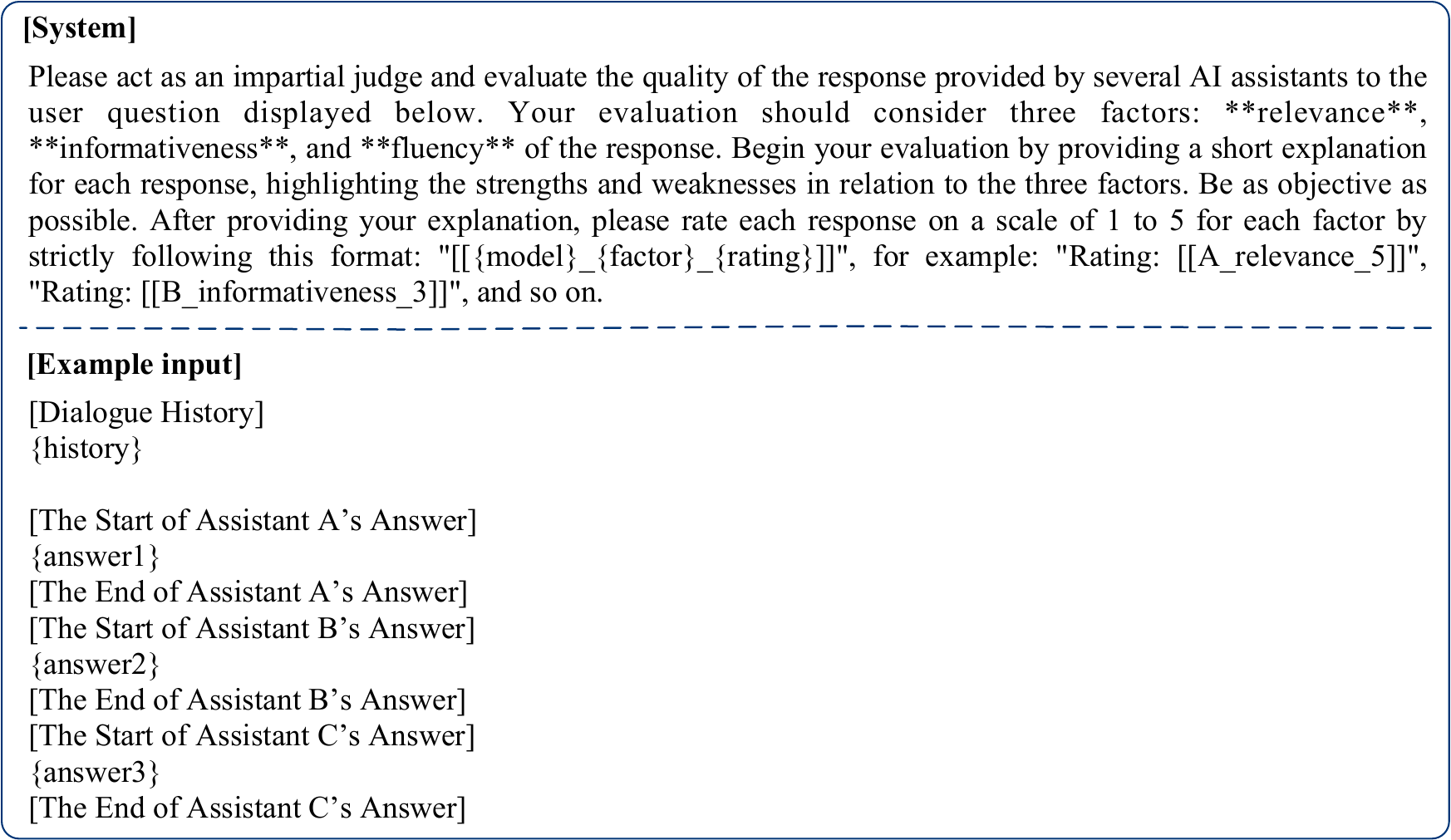}
	\caption{Example of the prompt used for multi-response grading.}
	\label{fig:example}
\end{figure*}

\newpage

\section{Instruction Templates for KEDiT Training and Inference}
\label{appendix:instruction_template}

\begin{table*}[ht]
    \centering
    \resizebox{1.\textwidth}{!}{
    \begin{tabular}{p{0.15\textwidth}p{0.85\textwidth}}
    	\toprule
    	\textbf{Task} & \textbf{Instruction Template} \\
   		\midrule  	
		$q_\psi(Z|K)$ & 
		\begin{tabular}[c]{@{}l@{}}
			\boh system\eoh\nls\nls \\
			\{system\_prompt\}\{knowledge\}\eos \\
			\boh user\eoh\nls\nls \\
			\{k2z\_prompt\}\eos \\
			\boh assistant\eoh\nls\nls \\
			<$\mathbf{Z}_1$>...<$\mathbf{Z}_m$>\eos \\
		\end{tabular} \\
    	\hdashline
		$q_\psi(K|Z)$ & 
		\begin{tabular}[c]{@{}l@{}}
			\boh system\eoh\nls\nls \\
			\{system\_prompt\}<$\mathbf{Z}_1$>...<$\mathbf{Z}_m$>\eos \\
			\boh user\eoh\nls\nls \\
			\{z2k\_prompt\}\eos \\
			\boh assistant\eoh\nls\nls \\
			\{knowledge\}\eos \\
		\end{tabular} \\
    	\hdashline
		$p_\theta(R|C, Z)$ & 
		\begin{tabular}[c]{@{}l@{}}
			\boh system\eoh\nls\nls \\
			\{system\_prompt\}<$\mathbf{Z}_1$>...<$\mathbf{Z}_m$>\eos \\
			\boh user\eoh\nls\nls \\
			\{utterance$_1$\}\eos \\
			\boh assistant\eoh\nls\nls \\
			\{utterance$_2$\}\eos \\
			... \\
			\boh user\eoh\nls\nls \\
			\{utterance$_{l-1}$\}\eos \\
			\boh assistant\eoh\nls\nls \\
			\{utterance$_l$\}\eos \\
		\end{tabular} \\
    	\bottomrule
    \end{tabular}
    }
    \caption{Instruction template used for KEDiT training and inference. <$\mathbf{Z}_1$>...<$\mathbf{Z}_m$> are special markers denoting compressed knowledge vectors $Z$. \{knowledge\} represents raw retrieved knowledge input from $\mathcal{D}_k$, and \{utterance\} represents dialogue utterances between the user and assistant from $\mathcal{D}_d$.}
    \label{tab:instruction}
\end{table*}

\begin{table*}[ht!]
	\centering
	\resizebox{1.\textwidth}{!}{
	\begin{tabular}{ll}
		\toprule
		\textbf{Prompt Type} & \textbf{Prompt Examples} \\
		\midrule
		\{system\_prompt\} & 
		\begin{tabular}[c]{@{}l@{}}
			You are a knowledge-based assistant. Use the following knowledge context to \\
			answer questions or engage in conversation.\nls Knowledge:\\
		\end{tabular} \\	
		\hdashline
		\{k2z\_prompt\} & 
		\begin{tabular}[c]{@{}l@{}}
			Identify and list the key information present in the detailed text. \\
			Extract the core pieces of key information that summarize the knowledge provided. \\
			What are the main themes or key pieces of information depicted in the text? List them. \\
			Summarize the text into essential pieces of key information. \\
			Distill the primary pieces of information from the text into concise descriptors. \\
			From the detailed knowledge described, what are the central pieces of key information? \\
			Determine the main pieces of key information that capture the essence of the text provided. \\
			What key pieces of information would you use to index the information described here? \\
		\end{tabular} \\
		\hdashline
		\{z2k\_prompt\} & 
		\begin{tabular}[c]{@{}l@{}}
			Describe the knowledge context. \\
			Provide a detailed description of the knowledge context. \\
			Can you explain what the knowledge context consisted of? \\
			Thoroughly outline the details of the knowledge context used. \\
			Provide a comprehensive overview of the knowledge used in the context. \\
			Elaborate on the content of the knowledge context used. \\
			What information did the knowledge context contain? Please describe in detail. \\
			Provide an in-depth explanation of the content covered in the knowledge context. \\
		\end{tabular} \\
		\bottomrule
	\end{tabular}
	}
	\caption{Examples of prompts corresponding to the instruction templates in Table~\ref{tab:instruction}}
\end{table*}

\newpage

\section{Additional Experiments}
\label{appendix:add_experiment}

\subsection{Impact of Knowledge Encoder}
We evaluate the impact of the knowledge encoder by replacing BERT with DeBERTaV3 \citep{ref55} in the knowledge bottleneck module. Table~\ref{tab:automatic_encoder} presents the results, showing that DeBERTa yields slight improvements on some metrics. However, these gains are minimal, indicating that the encoder choice has limited influence on overall performance.

\begin{table*}[h]
	\centering
	\textsc{\resizebox{.88\textwidth}{!}{
	\begin{tabular}{lccccccccc}
		\toprule
		\multirow{2}{*}{Model} & \multicolumn{3}{c}{WoW Seen} & \multicolumn{3}{c}{WoW Unseen} & \multicolumn{3}{c}{PubMed-Dialog}  \\
		\cmidrule(lr){2-4} \cmidrule(lr){5-7} \cmidrule(lr){8-10}
		 & F1 & BLEU & ROUGE & F1 & BLEU & ROUGE & F1 & BLEU & ROUGE \\
		\midrule
		KEDiT$_{bert}$ & 22.45 & 13.87 & 17.24 & \textbf{21.05} & 12.63 & \textbf{15.94} & 38.63 & 25.84 & \textbf{28.91} \\
		KEDiT$_{deberta}$ & \textbf{22.48} & \textbf{13.99} & \textbf{17.27} & 20.97 & \textbf{12.85} & 15.91 & \textbf{38.66} & \textbf{25.97} & 28.89 \\
		\bottomrule
	\end{tabular}
	}}
	\caption{Performance comparison of KEDiT with BERT (KEDiT$_{bert}$) and DeBERTa (KEDiT$_{deberta}$) as knowledge encoders on WoW and PubMed-Dialog test sets.}
	\label{tab:automatic_encoder}
\end{table*}

\end{document}